\documentclass{article}

\PassOptionsToPackage{numbers,compress}{natbib}
\usepackage[final,main]{neurips_2026}
\usepackage[utf8]{inputenc} %
\usepackage[T1]{fontenc}    %
\usepackage{hyperref}       %
\usepackage{url}            %
\usepackage{booktabs}       %
\usepackage{amsfonts}       %
\usepackage{nicefrac}       %
\usepackage{microtype} 
\usepackage{multirow}%
\usepackage{graphicx}
\usepackage{array}
\usepackage{makecell}
\usepackage{comment}
\usepackage{amsmath}
\usepackage{adjustbox}
\usepackage{graphicx}
\usepackage{wrapfig}
\usepackage{float}
\usepackage{pgffor}
\usepackage{subcaption}
\usepackage{adjustbox}
\usepackage{longtable}
\usepackage{booktabs}
\usepackage{placeins}
\usepackage{algorithm}
\usepackage[noend]{algpseudocode}
\usepackage{caption}
\usepackage{amsmath}
\usepackage{amssymb}
\usepackage{ltablex}
\usepackage{booktabs}
\usepackage{tabularx}
\usepackage{etoc}
\usepackage{titletoc}
\usepackage{makecell}
\usepackage{float} %
\usepackage{titlesec}
\usepackage{url}
\usepackage{appendix}
\usepackage[normalem]{ulem}
\keepXColumns
\usepackage{array} %
\usepackage[dvipsnames]{xcolor}
\usepackage[most]{tcolorbox}
\usepackage[all]{nowidow}
\usepackage[dvipsnames]{xcolor}

\newcommand{\newContent}[1]{{\color{blue} #1}\normalcolor}

\title{\textit{Retrieval Over Training} \\ Similarity search-based Model Selection for Time Series Anomaly Detection}

\author{
Christos Panourgias\textsuperscript{1,*},
Roberto Stanzione\textsuperscript{2,*},
Adrien Petralia\textsuperscript{1},
\\
\bfseries Themis Palpanas\textsuperscript{1},
Paul Boniol\textsuperscript{2}
\\[2pt]
\textsuperscript{1}Université Paris Cité, LIPADE
\qquad
\textsuperscript{2}École Normale Supérieure, Inria
\\[2pt]
\footnotesize
\texttt{\{panourgiaschris,adrien.petralia\}@gmail.com}
\qquad
\texttt{\{roberto.stanzione,paul.boniol\}@inria.fr}
\\[-1pt]
\footnotesize
\texttt{themis@mi.parisdescartes.fr}
\\[2pt]
{\small \textsuperscript{*}Equal Contribution}
}

\begin{document}

\maketitle

\begin{abstract}
    Anomaly detection is a fundamental task for time-series analytics, with important implications for the downstream performance of many applications. 
    Despite the large number of anomaly detection methods proposed in the literature, recent benchmark studies have shown that no single detector performs best across highly heterogeneous time series. 
    Therefore, a practical and scalable solution is to develop a model-selection method that, for a given time series, selects the anomaly detector most likely to perform well. 
    Nevertheless, the model selection approaches proposed in the literature suffer from a significant drop in accuracy when applied in Out-of-Distribution (OOD) settings.
    In this paper, we tackle the aforementioned limitation and propose \textbf{\textsc{RAMSAD}}, a retrieval-based framework for model selection in time-series anomaly detection.
    Given a new time series, our method queries a knowledge base of previously observed instances, retrieves the top-${k}$ most similar series, and transfers detector recommendations from their performance profiles. 
    The framework can operate with standard similarity measures as well as embedding-based representations. 
    Our results show that RAMSAD outperforms existing baselines in In-Distribution (ID) settings and is the most robust selector in OOD, where, when selecting a small subset of detectors, it matches or exceeds ensembles that run the entire detector pool, at a fraction of their cost. 
    Overall, we demonstrate that similarity-based retrieval constitutes a strong and efficient foundation for model selection in time-series anomaly detection.
\end{abstract}

\section{Introduction}

Time-series anomaly detection is a fundamental problem in machine learning, with applications in healthcare, manufacturing, finance, energy, cybersecurity, and scientific monitoring. 
Its goal is to identify observations or subsequences that deviate from expected temporal behavior. 
Because the utility of downstream monitoring systems depends directly on the quality of anomaly detection, selecting an appropriate detector is a critical practical challenge~\cite{boniol2024divetimeseriesanomalydetection,schmidl2022comprehensive}.

Despite substantial progress in the literature, time-series anomaly detection remains far from solved. 
A central difficulty is the extreme heterogeneity of time series encountered in practice. 
Real-world datasets differ widely in sampling rate, scale, length, noise characteristics, seasonality, dimensionality, and anomaly type.
Recent benchmark studies have reinforced this observation, showing that there is no universally best anomaly detector across heterogeneous collections of datasets~\cite{paparrizos2022tsb,liu2025tsb}. 
This variability makes \emph{model selection}
a key problem in its own right.

\setlength{\columnsep}{10pt} 

\begin{wrapfigure}{R}{0.47\linewidth}
  \centering
  \vspace{-2.5em}
  \includegraphics[width=1.0\linewidth]{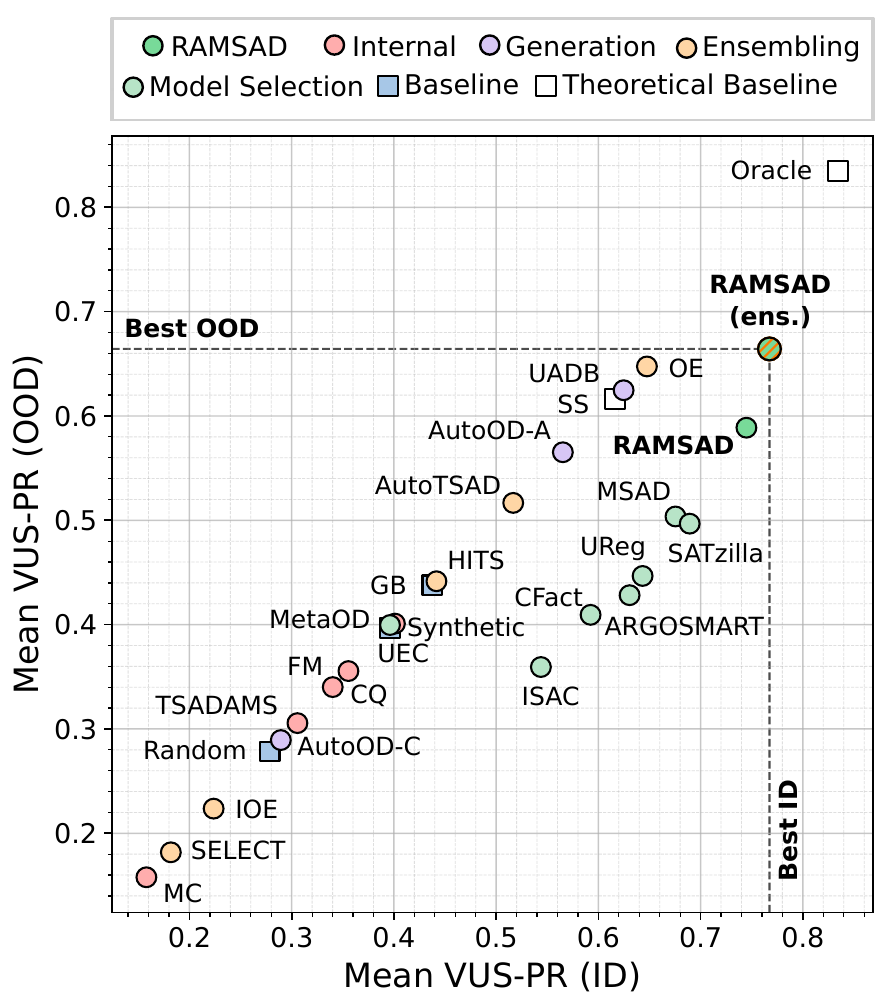}
  \caption{Comparison of RAMSAD with all automatic TSAD approaches (ID and OOD).
  }
  \label{fig:id_vs_ood}
  \vspace{-1em}
\end{wrapfigure}

Existing work has increasingly approached model selection with trainable strategies~\cite{sylligardos2023choose,sylligardosmsad2025}. 
These methods learn, from benchmark results or dataset representations, which detector to apply to a new series. While appealing, they add another layer of learning on top of an already difficult task, requiring additional supervision, tuning, and assumptions about generalization across domains. This complexity is not always justified, especially in heterogeneous settings where benchmark coverage is limited and domain shift is common. We therefore ask a simpler question: \emph{can anomaly-detector selection be solved 
without training a selector at all?}

\vspace{0em}

In this paper, we answer this question affirmatively. 
We propose \textbf{\textsc{RAMSAD}} (\underline{\textbf{R}}etrieval-\underline{\textbf{A}}ugmented \underline{\textbf{M}}odel \underline{\textbf{S}}election for \underline{\textbf{A}}nomaly \underline{\textbf{D}}etection), a retrieval-based framework for time-series anomaly detection based on a training-free, non-parametric selector, i.e., it does not require fitting a separate meta-learner; instead, it ranks candidate detectors directly from the performance of detectors on retrieved similar time series.
The idea is simple: similar time series are likely to favor the same anomaly detector.
We maintain a knowledge base of previously observed time series, together with the performance profiles of anomaly detectors when applied to these series; as for trained selectors, computing these profiles requires labels for the knowledge-base series, but never for the query.
Given a new query series, we retrieve its top-${k}$ nearest neighbors and use their outcomes to rank candidate detectors. 
This turns model selection into a similarity-search problem, eliminating the need for a dedicated trained selector.

Our framework is flexible with respect to the representation used for retrieval. 
In particular, it 
supports multimodal embeddings obtained by contrastive alignment of time series with textual dataset descriptions. 
We evaluate the proposed approach in both in-distribution (ID) and out-of-distribution (OOD) settings. 
In addition, as illustrated in Figure~\ref{fig:id_vs_ood}, selecting an ensemble of $N=10$ detectors improves robustness and achieves SOTA  in ID and OOD performance while remaining substantially faster than competing train-dependent selectors. 
Training-free refers to the selector only: (i)
like any supervised selector, RAMSAD needs a labeled knowledge base, but no labels for the query;
(ii) the optional alignment module is trained offline without detector identities or performance, and
queries need no description; and (iii) a frozen, off-the-shelf TSFM already outperforms all trained
selectors. Our main contributions are as follows:

\noindent$\bullet$ We introduce a retrieval-based formulation of anomaly-detector selection for time series, casting the problem as non-parametric inference over a knowledge base of prior instances.

\noindent$\bullet$ We introduce a \emph{segment-to-series} aggregation pipeline where 
we effectively translate segment-level similarity matches into a robust, series-level ranking.

\noindent$\bullet$ We propose \textsc{RAMSAD}\footnote{Code available at: \url{https://github.com/Hendrix8/RAMSAD}
} for both classical similarity measures and learned embedding spaces, including multimodal time-series--text representations. We experimentally show that aligning TSFM embeddings with simple dataset descriptions provides a useful semantic signal that mainly improves retrieval in OOD.

\noindent$\bullet$ We empirically demonstrate that \textsc{RAMSAD} is a strong alternative to trainable selectors, delivering higher accuracy in ID and greater robustness in OOD, while incurring substantially lower cost.

\section{Foundations and Related Work}
\label{sec:background_and_related}

\textbf{Notations.}
\label{sec:background_and_related_notations}
A dataset is denoted by \(\mathfrak{D}=\{T_1,\dots,T_{|\mathfrak{D}|}\}\), where each \(T_i\) is a univariate time series. Specifically, the \(i\)-th series is written as
\(
T_i = [x_{i,1}, x_{i,2}, \ldots, x_{i,L_i}] \in \mathbb{R}^{L_i},
\)
where \(L_i = |T_i|\) is the length of \(T_i\), and \(x_{i,t} \in \mathbb{R}\) denotes the value of \(T_i\) at time index \(t\). 
A subsequence of length \(\ell\) extracted from \(T_i\) and starting at position \(t\) is denoted by
\(
T_i^{(t,\ell)} = [x_{i,t}, x_{i,t+1}, \ldots, x_{i,t+\ell-1}] \in \mathbb{R}^{\ell}.
\)
The anomaly score of a detector \(D\) applied on \(T_i\) is denoted as
\(
S_i = [s_{i,1}, s_{i,2}, \ldots, s_{i,L_i}] \in \mathbb{R}^{L_i}.
\)

\textbf{Time-Series Anomaly Detection.}
\label{sec:background_and_related_tsad}
The latter can be grouped according to the underlying detection principle~\cite{boniol2024divetimeseriesanomalydetection}.
\textit{Distance-based} methods identify anomalous subsequences by comparing them against their nearest neighbors in the series or in a reference set~\cite{keogh2007compression,liu2009efficient,yeh2016matrix,linardi2020matrix}.
\textit{Density-based} methods instead detect anomalous behaviors by estimating how isolated observations or subsequences are in a representation space. This family includes \textit{distribution-based}~\cite{antoni2019conditionIndicators}, \textit{graph-based}~\cite{boniol2020series2graph,wenig2024series2graph++}, \textit{tree-based}~\cite{liu2008isolation, ma2020isolation}, and \textit{encoding-based}~\cite{Gao2020ensemble}.
Finally, \textit{prediction-based} methods estimate expected temporal behavior and use prediction error as the anomaly proxy. These approaches can be divided into \textit{forecasting-based}~\cite{Wu_2022} and \textit{reconstruction-based}~\cite{su2019robust}.
A detailed review is provided in Appendix~\ref{sec:appendix_review}.

{
\subsection{Automatic Solutions}
}

\label{sec:background_and_related_model_sel}

\begin{figure}[tb]
  \centering
  \includegraphics[width=1.0\linewidth]{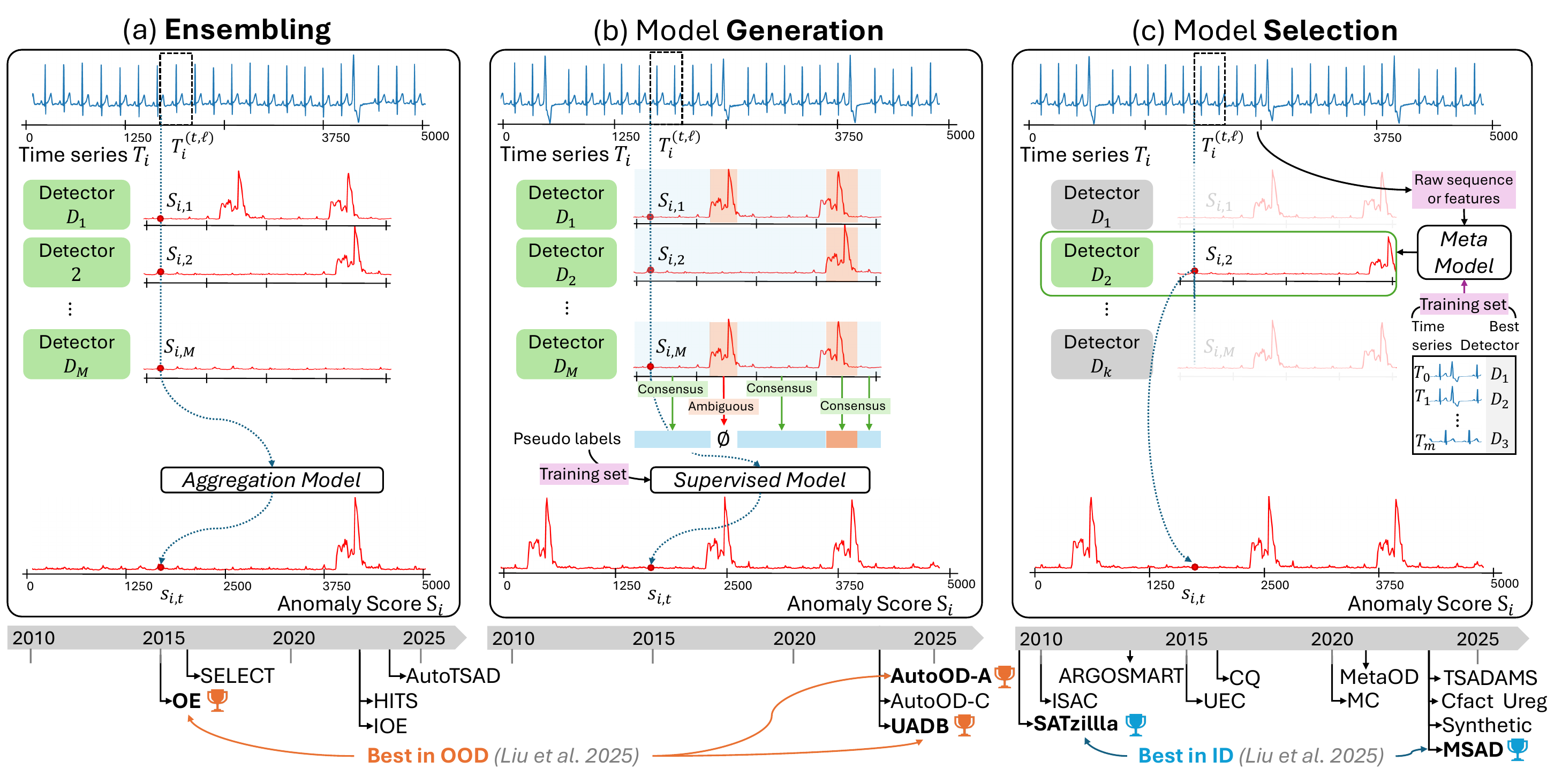}
  \vspace{-0.5cm}
  \caption{Automatic solutions for time series anomaly detection discussed in Section~\ref{sec:background_and_related_model_sel}.} 
  \vspace{-0.5cm}
  \label{fig:autoML}
\end{figure}

Recent benchmark studies for time-series anomaly detection %
~\cite{schmidl2022comprehensive,liu2024elephant} 
{have motivated automated strategies.} 
Existing approaches can be grouped into three main categories illustrated in Figure~\ref{fig:autoML}: \textit{model selection}, \textit{model ensembling}, and \textit{model generation}~\cite{liu2025tsb}. 

\textbf{Model Selection.}
Model selection methods aim to choose the best detector for a given time series.
A large subset of this literature follows the \textit{meta-learning} paradigm, where historical knowledge about detector performance is used to guide future choices.
Simple meta-learners such as \texttt{ARGOSMART}~\cite{nikolic2013simple} and \texttt{ISAC}~\cite{kadioglu2010isac} rely on similarity search over meta-features.
Other approaches formulate selection through,  (i) collaborative filtering
~\cite{zhao2021automatic}, (ii) as a classification problem~\cite{sylligardos2023choose,sylligardosmsad2025}, %
or (iii) as a regression problem that predicts expected detector performance scores~\cite{xu2008satzilla, navarro2023meta}.
A different line of work, denoted as \textit{Internal} in~\cite{liu2025tsb}, estimates detector quality directly on the target series without labeled historical data, using (i) unsupervised surrogate metrics computed from anomaly scores~\cite{goix2016evaluate}, (ii) clustering-based internal validation criteria~\cite{nguyen2016evaluation}, (iii) synthetic anomaly injection~\cite{chatterjee2022mospat},
or (iv) the Model Centrality~\cite{lin2020infogan}. 
Building on multiple surrogate criteria, \texttt{TSADAMS}~\cite{goswami2022unsupervised} aggregates detector rankings to obtain more robust recommendations.

\textbf{Model Ensembling.}
Model ensembling combines the outputs of multiple detectors in order to improve robustness and detection quality.
Outlier Ensemble (OE)~\cite{aggarwal2015theoretical} aggregates detector scores using simple operators such as the average, the maximum, or the average of maxima.
More selective variants first identify a subset of reliable detectors before aggregation~\cite{rayana2016less, ma2023need}.
Finally, \texttt{AutoTSAD}~\cite{schmidl2024autotsad} provides a fully automated pipeline that generates synthetic data, optimizes detector configurations, and ensembles the most effective detectors into a final anomaly score.

\textbf{Model Generation.}
Model generation 
typically relies on pseudo-labels that identify likely inliers and candidate outliers, and uses them to guide the combination of multiple detectors.
For example, \texttt{AutoOD-A}~\cite{cao2023autood} starts from a small set of pseudo-labels and iteratively expands it while pruning less effective detectors.
Conversely, \texttt{AutoOD-C}~\cite{cao2023autood} starts from a larger pseudo-labeled set and progressively removes labels.
Finally, \texttt{UADB}~\cite{ye2023uadb} combines information from multiple detectors into a single booster model that iteratively refines anomaly predictions.

{
\subsubsection{Limitations of Current Strategies: \textit{The scalability and OOD trade-off}}
}

\label{sec:background_and_related_limitations}

Existing automatic solutions for time-series anomaly detection involve a trade-off between robustness, flexibility, and computational efficiency. 
\textit{Model ensembling} and \textit{model generation} can improve robustness by combining information from multiple detectors, but they typically require running a large part---or the entirety---of the detector pool before producing a final anomaly score, \textbf{which can be prohibitively expensive for long time series or large detector collections (L1).}
\textit{Model selection} addresses this issue more directly by aiming to choose a single detector for each series, but most existing approaches rely on train-dependent classification or regression models. 
As a result, they are often tied to a fixed detector pool, \textbf{require retraining when the pool changes (L2)}, and may \textbf{generalize poorly under domain shift (L3)}. These limitations motivate the search for a simpler and more flexible alternative without requiring a dedicated training phase for model selection.

\subsection{Problem Definition}
\label{sec:background_and_related_problem}

Let \(\mathcal{B}=\{D_1,\dots,D_M\}\) be a finite pool of candidate anomaly detectors, and let \(\mathfrak{D}=\{T_1,\dots,T_{|\mathfrak{D}|}\}\) be a collection of time series. For a given series \(T_i \in \mathfrak{D}\) with ground-truth anomaly labels \(Y_i\), each detector \(D_j \in \mathcal{B}\) produces an anomaly score sequence \(S_{i,j}=D_j(T_i)\), whose quality is evaluated through a performance measure \(\mathrm{Acc}(S_{i,j},Y_i)\) (e.g., AUC-PR, F1, or VUS).
The \emph{model selection} problem is to construct a selector \(\Phi_K:\mathfrak{D}\rightarrow 2^{\mathcal{B}}\), which maps each time series \(T_i\) to a subset of \(K\) detectors, \(\Phi_K(T_i)\subseteq\mathcal{B}\). The case \(K=1\) recovers standard single-detector selection, while \(K>1\) corresponds to selecting an ensemble. Let \(g(\Phi_K(T_i),T_i)\) denote the score sequence obtained by applying the selected detector set to \(T_i\), using the ensemble aggregation rule when \(K>1\).
The goal is for this selector to match, as closely as possible, the \emph{oracle} choice
\(
\Phi_K^\star(T_i)=
\arg\max_{\mathcal{S}\subseteq\mathcal{B},\,|\mathcal{S}|=K}
\mathrm{Acc}\big(g(\mathcal{S},T_i),Y_i\big).
\)
Equivalently, model selection can be formulated as minimizing the expected regret:
\(
\mathbb{E}_{(T_i,Y_i)}\!\left[
 \mathrm{Acc}\big(g(\Phi_K^\star(T_i),T_i),Y_i\big)
- \mathrm{Acc}\big(g(\Phi_K(T_i),T_i),Y_i\big)
\right].
\)
Under this formulation, the objective is to predict which detector or detector subset from \(\mathcal{B}\) to select for a given series.

In this paper, we aim to address effectively model selection (addressing \textbf{L1}) using a framework based on a knowledge base and similarity search, and thus using a train-free selector (addressing \textbf{L2}).  The main challenge remaining is on the definition of the similarity space and the retrieval process that minimize the effect of domain drift (i.e., OOD) on accuracy (addressing \textbf{L3}).

{\textbf{Scope.} Throughout this paper, we consider static univariate time series, and OOD refers to \emph{domain shift}: the query series comes from a domain that is entirely absent from the knowledge base. Temporal concept drift within a domain, streaming settings, and non-stationarity within a single long series (e.g., an anomalous pattern becoming normal over time) are outside the scope of our evaluation; we discuss them in Appendix~\ref{app:limitations}.}

\section{RAMSAD: Proposed Approach}

\label{sec:proposed_approach}
\begin{figure}[tb]
  \centering
  \includegraphics[width=1.0\linewidth]{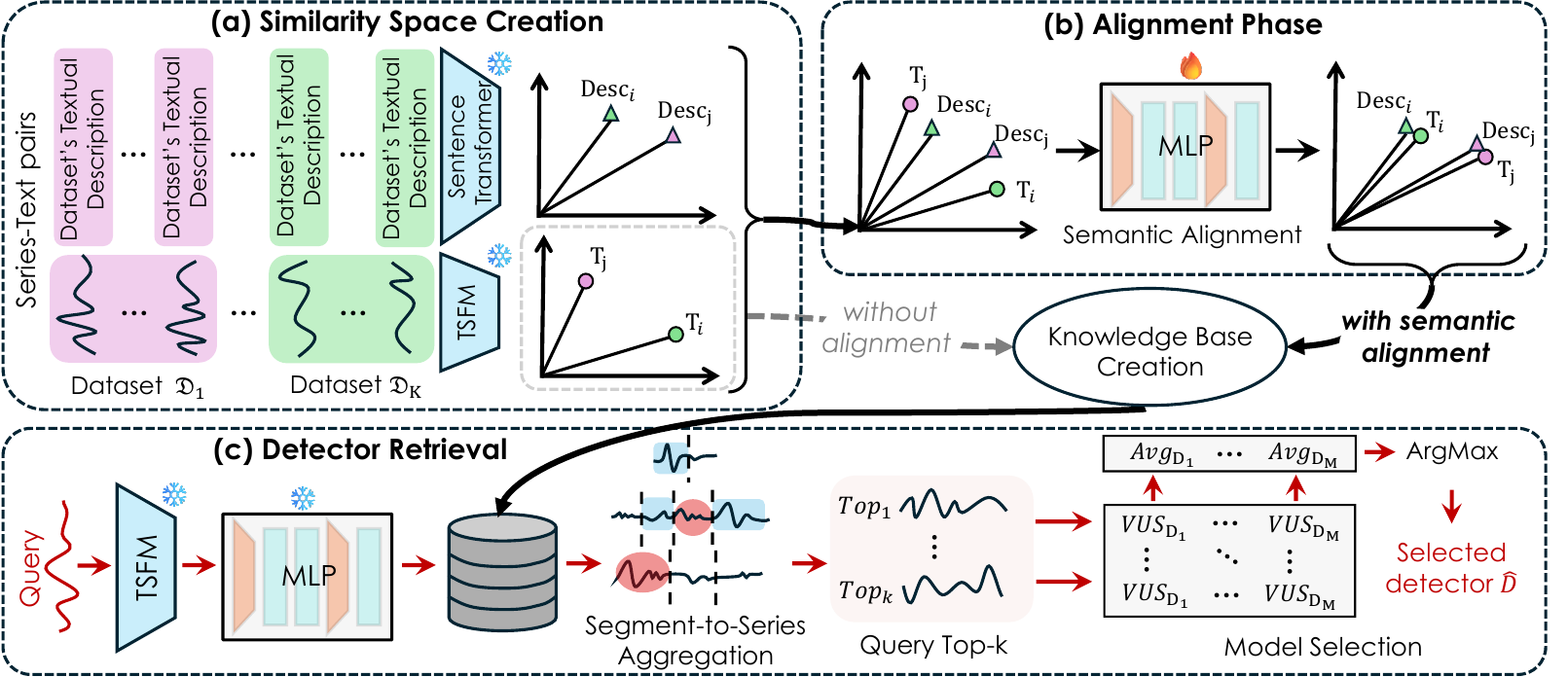}
  \vspace{-0.3cm}
  \caption{Overview of RAMSAD.} %
  \label{fig:framework}
  \vspace{-0.5cm}
\end{figure}

To address the limitations mentioned above, we propose \textbf{RAMSAD}, a retrieval-based framework for selecting anomaly detectors for time series (cf. Figure~\ref{fig:framework}).
Our approach relies on the following two core principles. %

\textbf{Similarity Space \& Alignment.}
From a time series database, we first partition each series into fixed-length segments and pair every segment with the textual description of its source dataset. We then encode the time-series segments with a TSFM to capture temporal structure, and encode the dataset descriptions with a sentence transformer to capture domain-level semantic information.
While we demonstrate in Section~\ref{sec:experiments_main_id} that our TSFM-based embedding is sufficient for in-distribution, the latter faces issues in OOD (i.e., Limitation \textbf{L3}).
To cope with \textbf{L3}, we align the two modalities (i.e., text and time series) with lightweight MLP module that learns a shared vector space in which temporally and semantically related items are mapped close to one another. The aligned segment representations are then stored in a knowledge base together with 
the known accuracy scores across the candidate detectors.
{This semantically informed space is particularly useful in anomaly detection, where the interpretation of anomalous behavior depends strongly on context.} %
For example, a sudden spike may indicate a fault in one domain but correspond to normal peak activity in another.
{By incorporating semantic information, the framework can better capture such differences and thus provide more reliable neighbors for model selection, especially in OOD settings.}
{Alignment is an optional, offline refinement: the textual descriptions only shape the time-series space during training, so neither the knowledge base nor the query requires a description at inference time.}

\textbf{Detector Retrieval.}
At inference time, a query series is processed through the same segmentation and encoding pipeline and projected into the shared space {using only the time-series projection head}. Retrieval is first performed at the segment level, where the nearest neighbors of the query segments are identified.
{They are then aggregated into the top-\(k\) retrieved series.} For these retrieved series, we access the %
precomputed VUS matrix and compute, for each candidate detector, the average VUS over the retrieved neighbors. The detector with the highest average VUS is then selected. %

{ 
\subsection{The Core Retrieval Pipeline}
}
\label{sec:core_retrieval}
We now describe the core retrieval mechanism of RAMSAD. %
The same pipeline can operate on the original space, TSFM embeddings, or our aligned representations presented in Section~\ref{sec:semantic_alignment}.

\textbf{Similarity Space Creation}.
RAMSAD is agnostic to the underlying similarity space. In the simplest setting, retrieval may be performed directly in the original time-series space, in which case segmentation is not necessary. For instance, Euclidean similarity can be computed on full series after truncating or padding them to a common length. 
We study such original space and distance alternatives in Appendix~\ref{app:similarity_spaces}.

Our main instantiation instead relies on embeddings produced by a TSFM~\cite{ansari2025chronos2}.
We use TSFMs because they capture temporal structure more effectively than raw pointwise distances and standardize heterogeneous time series into fixed-dimensional representations. To build this embedding space, we partition each series into fixed-length segments of size \(\min(L_i,8192)\), where \(8192\) is the maximum input size supported by \texttt{Chronos2}~\cite{ansari2025chronos2} in one of our instantiations. Each segment is encoded as
\(
z_{i,t}^{\mathrm{ts}} = f_{\mathrm{ts}}(T_i^{(t,\ell)}),
\)
with \(z_{i,t}^{\mathrm{ts}} \in \mathbb{R}^d\), where \(f_{\mathrm{ts}}\) may be instantiated with any TSFM or other time-series embedding backbone.
{Segmentation is therefore required by the bounded context of the encoder: it maps series of very different lengths to fixed-dimensional vectors that can be compared directly, and the aggregation step rewards candidate series that match the query consistently and accurately. 
We do not resample series and therefore make no claim of invariance to the sampling rate; segmentation may also discard dependencies longer than the encoder context (Appendix~\ref{app:limitations}).}

\textbf{From Segment-to-Series Aggregation}.
Retrieval in RAMSAD is performed at the \emph{segment} level, whereas model selection is defined at the \emph{series} level. We therefore introduce an aggregation step that converts segment-level matches into a ranked list of candidate series. For simplicity, we use \(k\) throughout to denote the retrieval size; because different query segments may retrieve segments originating from the same series, the final number of distinct candidate series may be smaller than \(k\).

Let \(T_q\) denote a query time series, and let \(\{T_q^{(t_u,\ell)}\}_{u=1}^m\) be the \(m\) extracted segments. For each segment $T_q^{(t_u,\ell)}$, we retrieve its top-\(k\) nearest segments, \(\{T_{j_{u,r}}^{(t_{u,r},\ell)}\}_{r=1}^k\), together with their corresponding distances
\(
d_{u,r}=d(T_q^{(t_u,\ell)},\,T_{j_{u,r}}^{(t_{u,r},\ell)}),
\)
for \(r\in\{1,\dots,k\}\).
Each retrieved segment \(T_{j_{u,r}}^{(t_{u,r},\ell)}\) originates from the indexed time series \(T_{j_{u,r}} \in \mathfrak{D}\). For each candidate series \(T_j \in \mathfrak{D}\), we collect all retrieved distances associated with segments originating from \(T_j\):
\(
\mathcal{A}(T_j)=\{\,d_{u,r} \;:\; j_{u,r}=j\,\}.
\)
We then define a base similarity score, where smaller values indicate stronger similarity, as
\(
\mathrm{base}(T_j)=\frac{1}{|\mathcal{A}(T_j)|}\sum_{d\in\mathcal{A}(T_j)} d.
\)
\begin{wrapfigure}[15]{r}{0.32\textwidth} 
  \vspace{-1em}
  \centering
  \includegraphics[width=1.0\linewidth]{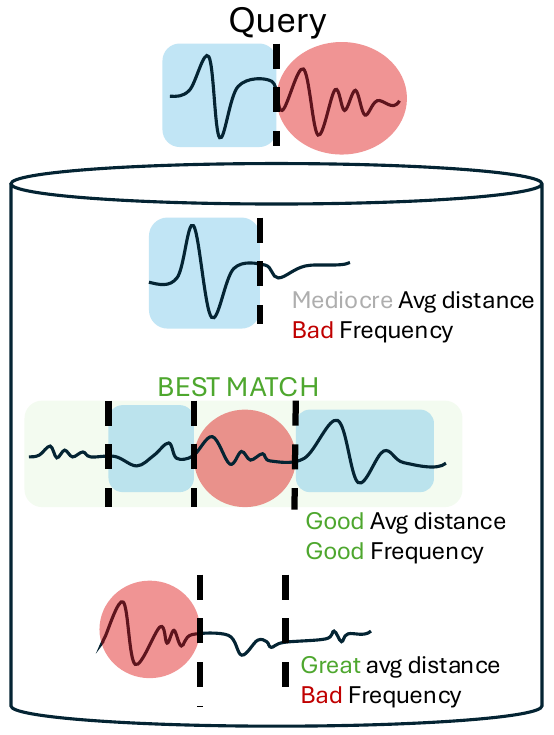}
  \caption{Segment-to-series aggregation in RAMSAD. %
  }
  \label{fig:retrieval}
\end{wrapfigure}
\vspace{1em}
A candidate series may be retrieved by many query segments or only by a few isolated ones. To distinguish consistent support from sporadic matches, we define the frequency of a candidate series \(T_j\) as the fraction of query segments for which at least one of the top-\(k\) retrieved neighbors originates from \(T_j\):
 \(
    \mathrm{freq}(T_j)
    =
    \frac{1}{m}
    \Bigl|
    \{\,u \in \{1,\dots,m\} : \exists r \in \{1,\dots,k\} \text{ s.t. } j_{u,r}=j \,\}
    \Bigr|
    \)
We then combine $\mathrm{freq}$ with the base similarity score to obtain the final ranking score:
\(
\mathrm{score}(T_j)
=
\frac{1}{\max(\epsilon,\mathrm{freq}(T_j))}\cdot \mathrm{base}(T_j),
\)
where \(\epsilon>0\) is a small constant. Candidate series are ranked in ascending order of \(\mathrm{score}(T_j)\), and the top-\(k\) retrieved series for query \(T_q\) are given by
\(
\mathrm{top_k}(T_q)
=
\arg\min_{T_j \in \mathfrak{D}}^{k}\ \mathrm{score}(T_j).
\)

Averaging distances across retrieved segments reduces sensitivity to isolated mismatches, while the frequency-aware term favors candidate series that receive consistent support across multiple query segments. 
The intuition of this approach is shown in Figure~\ref{fig:retrieval}.

\textbf{Model Selection}.
Once the top-\(k\) most similar series are retrieved, RAMSAD transfers detector recommendations from their stored performance profiles.
We assume access to a precomputed performance table
\(
\mathbf{V} \in \mathbb{R}^{|\mathfrak{D}| \times |\mathcal{B}|},
\)
where \(\mathbf{V}(T,D)\) denotes the accuracy (e.g., VUS-PR) of detector \(D \in \mathcal{B}\) on series \(T \in \mathfrak{D}\).
{Computing \(\mathbf{V}\) requires a labeled knowledge base on which every candidate detector has been evaluated; this is the same supervision consumed by trained meta-learning selectors, whereas no label is needed for the query.}
For each detector we aggregate its performance over the retrieved neighbors:
{\(
\bar{V}_D(T_q)
=
\frac{1}{|\mathrm{top}_k(T_q)|}\sum_{T \in \mathrm{top}_k(T_q)}\mathbf{V}(T,D).
\)}
The selected detector for query \(T_q\) is then
\(
\widehat{D}(T_q)
=
\arg\max_{D \in \mathcal{B}} \bar{V}_D(T_q).
\)
{When selecting an ensemble of \(N>1\) detectors, the \(N\) detectors with the highest \(\bar{V}_D(T_q)\) are selected.
Because the full performance profile of all candidate detectors is averaged over the neighborhood, a detector is selected only if it is broadly strong across the retrieved series.
Similar series may favor different detectors, and the retrieved neighbors therefore need not agree on their individual best detector.

{
\subsection{Enhancing the Similarity Space: Multimodal Semantic Alignment}
\label{sec:semantic_alignment}
}
We now describe the semantic alignment module that incorporates dataset-level context. 
As illustrated in Figure~\ref{fig:alignment}, semantic alignment structures TSFM embeddings using dataset-level context.
This motivates aligning each segment with both its paired description and related descriptions.

\begin{wrapfigure}[19]{R}{0.35\linewidth}
  \vspace{-1.6em}
  \centering
  \includegraphics[width=0.9\linewidth]{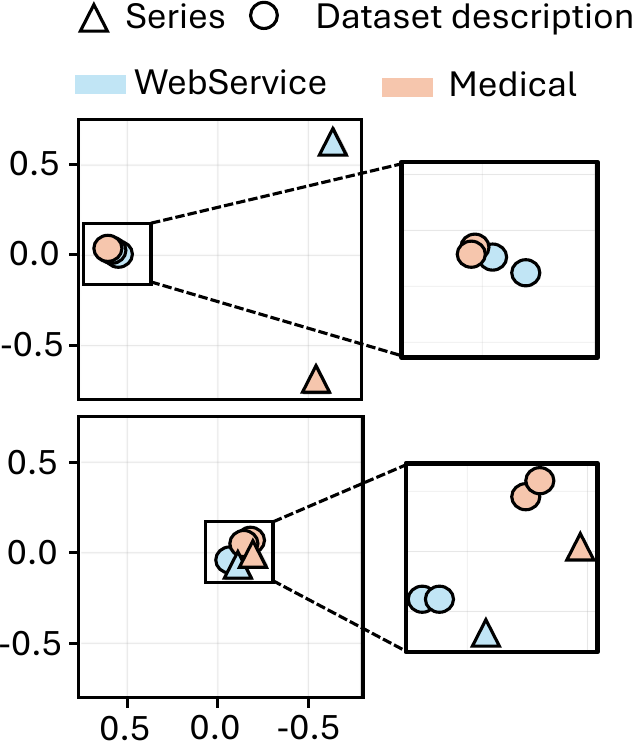}
  \caption{Before alignment (top), TSFM embeddings lack semantic structure; semantic alignment injects dataset-level description information (bottom).}
  \label{fig:alignment}
  \vspace{-2em}
\end{wrapfigure}

\textbf{Integrating Dataset Descriptions}.
Each segment \(T_i^{(t,\ell)}\) inherits the textual description of the dataset from which it is drawn. Let \(c_i\) denote the description associated with the dataset of series \(T_i\), and let
\(
z_i^{\mathrm{desc}} = f_{\mathrm{text}}(c_i)
\)
be its text embedding, where \(z_i^{\mathrm{desc}} \in \mathbb{R}^d\). In practice, \(f_{\mathrm{text}}\) is instantiated with the \texttt{all-mpnet-base-v2} Sentence Transformer. 
Since different datasets may belong to related domains, their descriptions can be similar even when the corresponding time-series segments are not identical.

\textbf{Description Consistency and Multi-Positive Matching}.
A one-to-one alignment between time-series segments and dataset descriptions is overly restrictive, since different datasets often carry highly similar textual descriptions. {In standard contrastive training, each segment has exactly one positive description, and every other description in the batch is treated as a negative and pushed away. 
When two datasets have near-synonymous descriptions then the description of the other dataset is wrongly declared a negative, which injects label noise in proportion to the number of near-
duplicate descriptions in the benchmark}.
To account for this, we compute a description-similarity matrix from pairwise dot products between text embeddings. 
We then threshold this matrix at \(\gamma=0.95\) and treat any pair of descriptions with similarity \(\geq\gamma\) as semantically compatible.

This yields a binary \emph{multi-positive} matching matrix \(P \in \{0,1\}^{B\times B}\) over a batch of \(B\) time-series-segment--description pairs. An entry \(P_{ij}=1\) indicates that the \(j\)-th text description is considered a valid positive for the \(i\)-th segment, i.e., the segment may plausibly originate from the domain or context described by that text. 
This matrix can therefore contain multiple positive entries per row. 
We additionally mask out in-batch pairs coming from the same time series, so that valid positives must originate from a different series than the anchor.

\textbf{Symmetric Alignment Objective}.
Because one description may match multiple segments and one segment may be compatible with multiple descriptions, we optimize a symmetric multi-positive contrastive objective. 
The objective is \emph{symmetric} because it is applied in both directions: from time-series segments to descriptions and from descriptions to time-series segments. 
The exact formulation of the loss is given in Appendix~\ref{app:semantic_alignment}.

\textbf{Using the Aligned Space at Inference}.
The alignment module has two projection heads, one per modality (Appendix~\ref{app:semantic_alignment}), and the text head is only used inside the training objective. 
At knowledge-base construction and at query time, all series follow the same path 
and retrieval is entirely time-series-to-time-series: text embeddings are neither stored in the index nor computed for a query, so a query never requires a description.
Moreover, the alignment objective only uses the correspondence between segments and descriptions; it never uses detector identities or detector-performance values, so it is independent of the selection task, and the selector itself remains training-free.
In our OOD experiments, the alignment module is trained without the target domain and is applied to that domain without any retraining: a query from an unseen domain is mapped close to the semantically closest domains covered by the knowledge base.

{
\subsection{System Complexity}
}

In RAMSAD, the dominant cost arises from segment retrieval. With exact search, processing a batch of \(b\) query segments requires computing similarities against all \(n\) indexed segments, leading to \(\mathcal{O}(bnd)\) query time and \(\mathcal{O}(nd +|\mathfrak{D}|M)\) persistent storage for the embedding database and the performance matrix. 
{An HNSW index avoids exhaustive comparisons at a small extra storage cost,} 
{making retrieval sublinear.} 
A detailed analysis is deferred to Appendix~\ref{app:system_complexity}.

\section{Experimental Evaluation}
\label{sec:experimental_evaluation}
We first compare our proposed approach with SOTA automated solutions both in terms of accuracy and efficiency. We then evaluate it when using different similarity spaces.
We provide additional analysis for scalability (Appendix~\ref{app:system_complexity}) and parameter sensitivity (Appendix~\ref{app:similarity_spaces}).

\textbf{Experimental Setup:} We evaluate on TSB-AD~\cite{liu2024elephant}, which includes 870 univariate time series from 9 domains, using the same train/test splits as recent model selection benchmarks~\cite{liu2025tsb} (details in Appendix~\ref{app:datasets}).
We compare against 25 automated solutions, including model selection, ensemble, model generation, and simple reference strategies such as Oracle, SS, GB, Random, and FM; all methods are described in Appendix~\ref{app:baselines}.
{Among these, Oracle and Supervised Selection (SS) are not deployable methods but oracle-style upper-bound references: Oracle picks the best detector of each test series, while SS selects the detector using the labels of a portion of the target series itself, which are unavailable to RAMSAD and to all other selectors.} 
The candidate pool contains 32 univariate time-series anomaly detectors from TSB-AD~\cite{paparrizos2022tsb}, listed in Appendix~\ref{app:detectors} {where 19 of them are also applicable to multivariate series~\cite{liu2025tsb}.}
We evaluate accuracy using VUS-PR~\cite{paparrizos2022volume,boniol2025vus}.
For RAMSAD, we test $k \in [1,10]$, and use $k=6$ as default. {In ID, the knowledge base holds the $619$ training series as $3{,}264$ segments, and the $251$ test series give $1{,}383$ query segments. In OOD, we hold out each of the $9$ domains in turn, removing its series from the knowledge base and from alignment training such that no query has a same-domain neighbor, and no retraining is performed for the target domain.}

\begin{figure}[tb]
\hspace{0cm}
  \centering
  \includegraphics[trim={0cm 0cm 0cm 0cm},clip,width=1.0\linewidth]{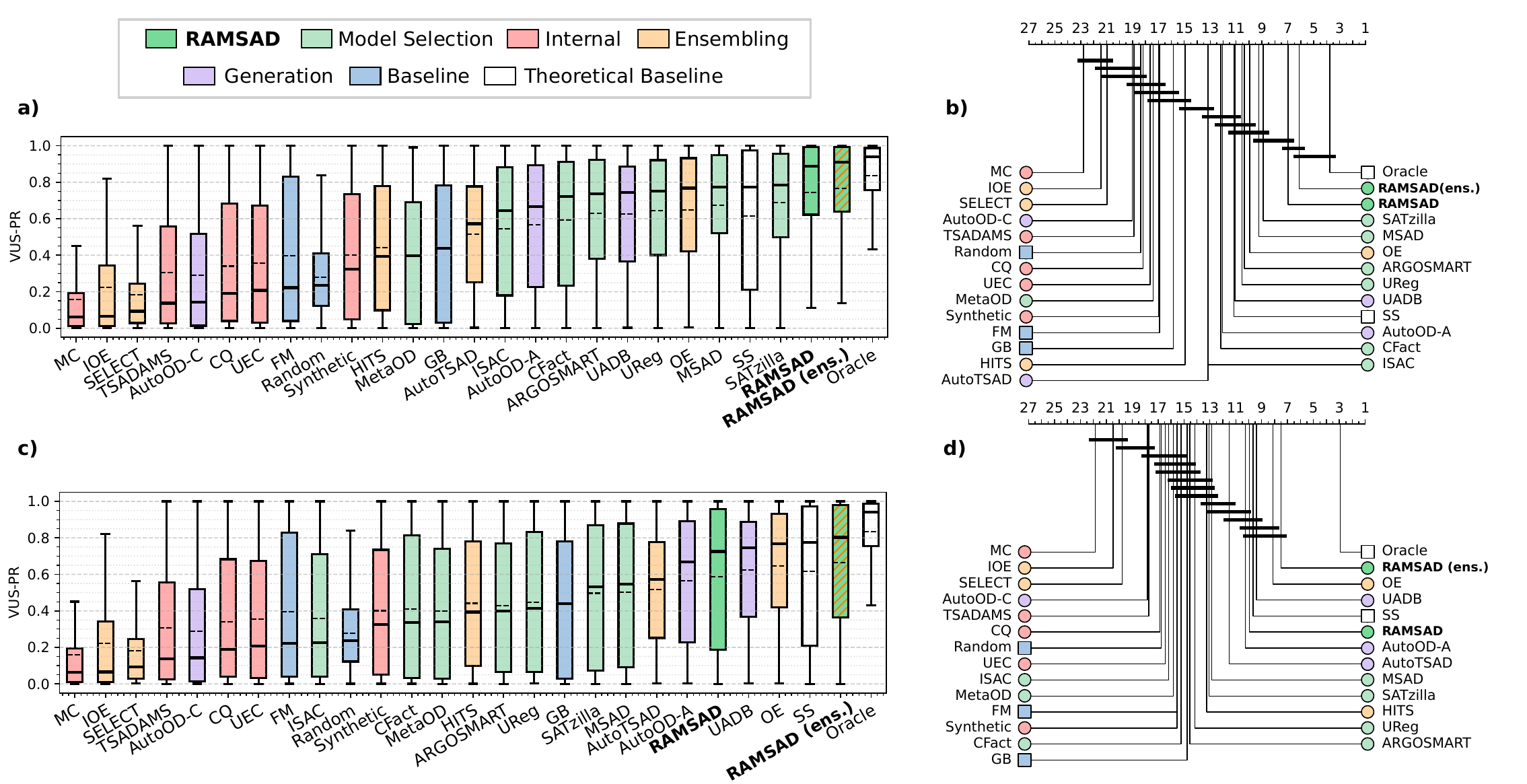}
  \vspace{-0.3cm}
  \caption{VUS-PR comparison for in-distribution (panel a) and out-of-distribution settings (panel c), with the associated Critical Difference Diagrams (panels b and d, \textit{resp.}).}
  \vspace{-0.5cm}
  \label{fig:boxplots}
\end{figure}

\textbf{In-Distribution Evaluation (ID)}. \label{sec:experiments_main_id}
Figure~\ref{fig:boxplots} (a) shows the overall VUS-PR score on the in-distribution test split (i.e., $251$ series) for all approaches, which are categorized according to the underlying methodology. 
{RAMSAD outperforms} all other baselines, achieving an average VUS-PR score of 
$0.74$. %
 {This shows that a retrieval-based selector, without any trained meta-learner, can outperform trained meta-learning selectors.}
Among other baselines, the best model selector is SATZilla, with an average VUS-PR of $0.69$; 
the best ensemble approach is OE, with an average VUS-PR of $0.64$; the best model generation is UADB with an average VUS-PR of $0.62$.
Unsupervised model selection methods lag significantly behind; the best performer, Synthetic~\cite{chatterjee2022mospat}, achieves only $0.40$, highlighting the difficulty of unsupervised selection.
{In the critical difference diagram (Figure~\ref{fig:boxplots}(b)),}
RAMSAD stands out as the best overall approach with an average rank of $6.97$,{ ahead of} the best baseline, SATZilla, which recorded an average rank of $8.90$.

\begin{wrapfigure}{24}{0.50\linewidth}
  \vspace{-0.8em}
  \centering
  \includegraphics[width=\linewidth]{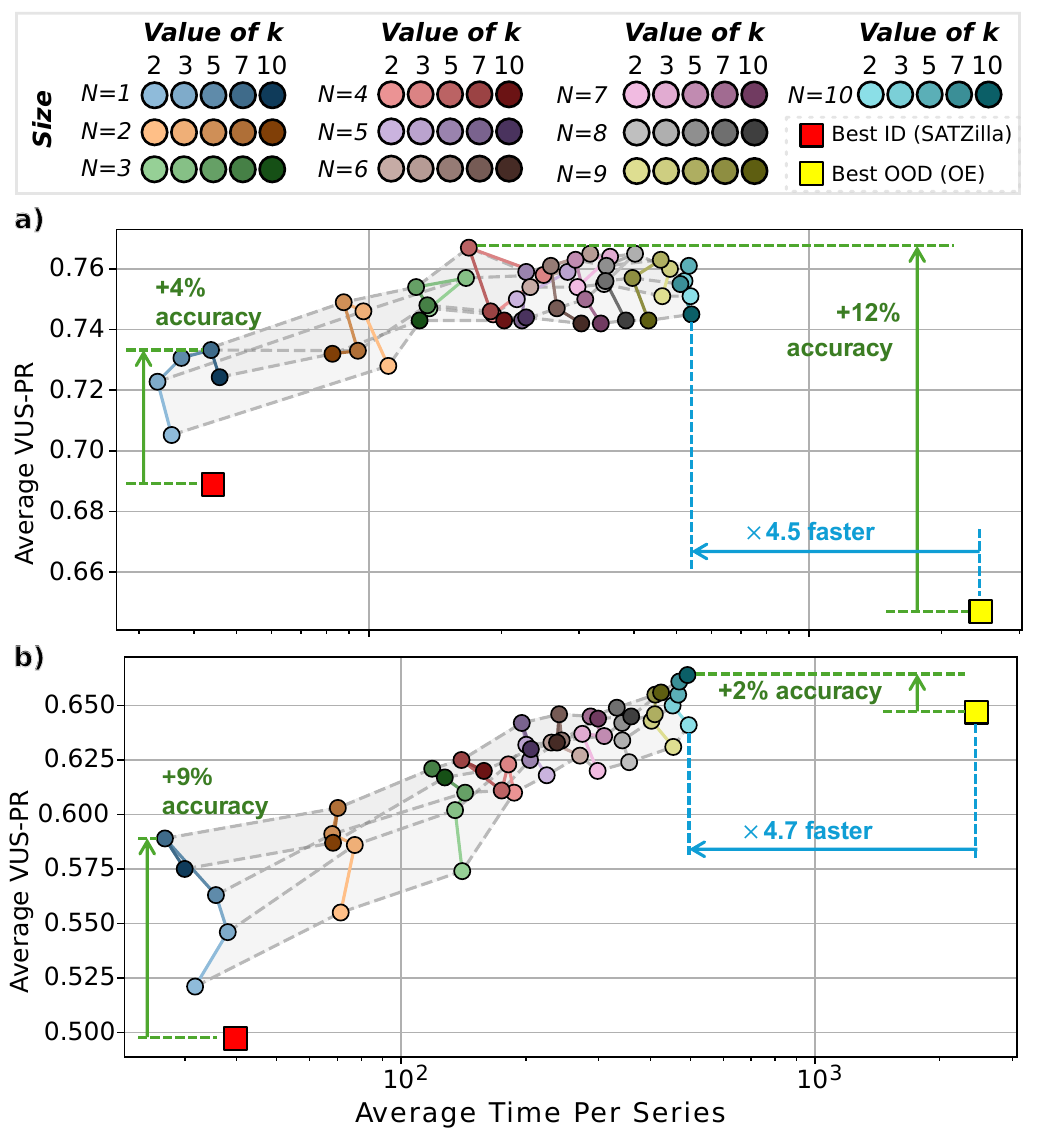}
  \caption{RAMSAD anomaly detection Runtime vs Accuracy by varying $k$ and $N$.}
  \label{fig:ensemble_id_ood}
  \vspace{-1em}
\end{wrapfigure}
\textbf{Out-of-Distribution Evaluation (OOD).}
We now consider the out-of-distribution scenario, in which for a time series of a given domain, model selectors can only consider a training set composed of time series from other domains. 
This can be noticed in Figure~\ref{fig:boxplots} (c): all model selection approaches are affected by OOD{, which is expected since every selector
must generalize to a domain that is absent from
its training data}. For example, SATZilla has an average VUS-PR of $0.50$ ($-0.19$ w.r.t. the ID setting), and in general all selection approaches showed VUS-PR drops of around $0.19$--$0.20$. RAMSAD is also challenged by OOD, achieving an average VUS-PR of $0.59$. However, RAMSAD is less affected than other selectors, with a drop of $-0.16$, showing better robustness and even managing to outperform some of ensembles and model generation approaches such as AutoTSAD and AutoOD-A. %
The CD plot in Figure~\ref{fig:boxplots} (d) summarizes the average ranking of each approach, with OE achieving the best ranking ($7.5$), followed by UADB ($8.7$). Notably, the performance of RAMSAD (avg. rank of $9.9$) is not statistically different from OE and UADB, widening the gap with other model selectors that were closer in the ID scenario. 
{RAMSAD therefore attains comparable robustness at a fraction of the cost, matching these methods while running a single detector.}
{Despite being the best approaches in the OOD setting, OE and UADB need to run all the $32$ detectors. However, we will show in the following section that RAMSAD can outperform these approaches by selecting a much smaller subset of detectors.}

\begin{wrapfigure}[14]{R}{0.48\textwidth}
  \vspace{-1.2em}
  \centering
  \includegraphics[width=\linewidth]{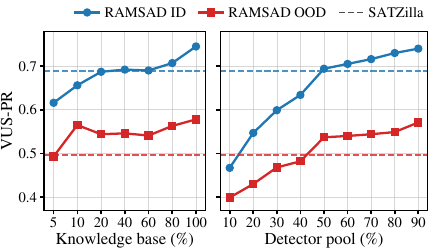}
  \caption{\textsc{RAMSAD} VUS-PR when randomly reducing the knowledge base (left) and the detector pool (right).}
  \label{fig:reduced_sizes_main_txt}
  \vspace{-1em}
\end{wrapfigure}
\textbf{Selecting Multiple Detectors}.
\label{sec:experiments_main_ensemble}
Figure~\ref{fig:ensemble_id_ood}(a) shows the average VUS-PR achieved by \textsc{RAMSAD} in ID scenarios when selecting ensembles of $2$ to $10$ detectors, illustrating the effect of different values of $k$ and comparing against the best ID and OOD baselines, SATZilla and OE, respectively, while also reporting the execution time required to run the selected detector(s). 
Already with a single detector ($N=1$), \textsc{RAMSAD} outperforms SATZilla by $4$\%. As a further benefit, selecting ensembles increases the average VUS-PR up to $0.77$, widening the gap to SATZilla to $8$\% and achieving a $12$\% accuracy improvement over OE.
In OOD (Figure~\ref{fig:ensemble_id_ood}(b)), \textsc{RAMSAD} outperforms SATZilla by $9$\% with $N=1$, and accuracy keeps improving as $N$ increases: with $N=10$ the gap to SATZilla grows to $17$\%, and \textsc{RAMSAD} exceeds OE by $2$\% while being $\times4.7$ faster. The latter is further supported by the boxplots and CD diagrams in Figure~\ref{fig:boxplots}. The ensemble version of \textsc{RAMSAD} outperforms all baselines in ID and OOD, while being one order of magnitude faster than the second best in OOD. Most notably, \textsc{RAMSAD} ensemble has no statistically significant difference from the oracle.

\label{sec:experiments_parameter}
\textbf{Parameter Sensitivity}.
Figure~\ref{fig:ensemble_id_ood} shows the behaviour of \textsc{RAMSAD} for different values of $k$ and $N$ in both ID and OOD. 
Overall, \textsc{RAMSAD} is stable across values of $k$: in the ID setting, VUS-PR varies only marginally, with a maximum difference of about $0.04$, while in the OOD setting the variation is slightly larger, up to $0.07$ for $N=1$, but decreases as the ensemble size increases; in all cases, \textsc{RAMSAD} consistently outperforms the best model selector, SATZilla.

\newContent{
\begin{table}[tb]
  \centering
  \small
  \caption{Latency from query series to detector recommendation, for single-detector selectors}
  \label{tab:selection_latency}
  \setlength{\tabcolsep}{4pt}
  \begin{tabular}{@{}lcccccccc@{}}
    \toprule
    & \textbf{RAMSAD} & ARGOSMART & ISAC & MSAD & SATZilla & UReg & MetaOD & CFact \\
    \midrule
    Features  & \textbf{19.6} & \multicolumn{7}{c}{37.6 (catch22, shared)} \\
    Inference & 0.05 & 1.1 & 25.6 & 25.8 & 39.7 & 221.8 & 426.7 & 453.4 \\
    \midrule
    Median    & \textbf{19.6} & 38.8 & 63.3 & 63.6 & 77.1 & 258.0 & 463.4 & 489.6 \\
    Std       & \textbf{0.4} & 7.8 & 7.8 & 8.3 & 12.2 & 25.1 & 28.6 & 32.2 \\
    \bottomrule
  \end{tabular}
\end{table}
}

\textbf{Selection Latency}.
\label{sec:experiments_latency}
Beyond the cost of running the selected detectors, Table~\ref{tab:selection_latency} isolates the cost of the selection itself, from a raw query series to a recommendation. \textsc{RAMSAD} is the fastest selector ($20$\,ms), where its TSFM forward pass is cheaper than the catch22 extraction used by the meta-learners, which alone takes $38$\,ms. RAMSAD is also the most predictable as its latency varies by 0.4 ms across queries.
End-to-end, single-detector selectors cost one detector run plus $39$--$490$\,ms, \textsc{RAMSAD} one run plus $20$\,ms ($N=1$) or ten runs plus $20$\,ms ($N=10$), and full-pool methods $32$ runs. As OE is the cheapest of the latter, the $\times4.7$ speed-up of \textsc{RAMSAD} ($N=10$) over OE lower-bounds its speed-up over UADB, AutoTSAD, AutoOD-A, and HITS.

\textbf{Knowledge-Base Coverage and Detector-Pool Size}.
\label{sec:experiments_kb}
Since RAMSAD relies on a labeled knowledge base, we study how its coverage affects selection (Figure~\ref{fig:reduced_sizes_main_txt}; details in Appendix~\ref{app:ablations}). Reducing the knowledge base, RAMSAD still outperforms SATZilla in OOD with $10\%$ of the training set, and in ID is comparable from $20\%$ and better from $80\%$. Reducing the detector pool, it matches or outperforms SATZilla with half of it.
With the leave-one-domain-out protocol this indicates that what matters is proximity in the representation space rather than shared domain membership.

\textbf{{Similarity Spaces Comparison}}.
We compare three retrieval spaces for RAMSAD: (1) original time-series, (2) TSFM embedding, and (3) semantically aligned.

Figure~\ref{fig:tsfm_boxplots} summarizes the main trends (full evaluation in Appendix~\ref{app:similarity_spaces}).
First, DTW, the strongest original-space baseline, reaching VUS-PR of \(0.69\) in ID and \(0.44\) in OOD, remains below the best TSFM and aligned-TSFM spaces. By comparison, retrieval in the plain TSFM, such as \texttt{Chronos2}, reaches \(0.75\) in ID and \(0.59\) in OOD.
{Second, the strongest overall behavior is obtained in the aligned TSFM space.} 
Figure~\ref{fig:tsfm_boxplots} shows that the aligned-TSFM performances are similar to the plain TSFM space in ID, and{ improve on them in OOD.} 
In OOD, the domain of the query is unknown, and temporal patterns alone may be insufficient to identify the most informative source series. 
In this case, incorporating semantic information becomes more important, which explains why the aligned space yields a clearer advantage in OOD.
{Dataset-level descriptions inject a coarse domain prior that, for some backbones, slightly lowers ID accuracy while improving cross-domain transfer  (Appendix~\ref{app:id_ood_tables}). Alignment is thus an OOD-oriented refinement rather than a prerequisite.

\begin{figure}[t]
  \centering
  \begin{minipage}[b]{0.48\textwidth}
    \centering
    \includegraphics[width=\linewidth]{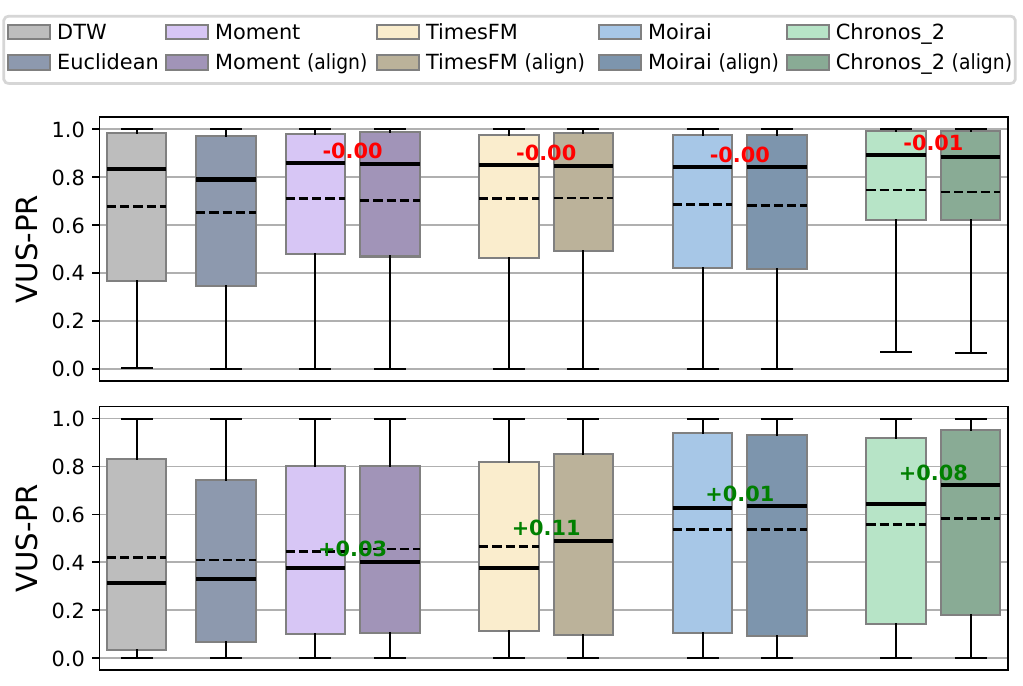}
    \caption{Comparison between original space (Euclidean and DTW), TSFM (light color), and aligned TSFM embedding spaces (darker color).}
    \label{fig:tsfm_boxplots}
  \end{minipage}\hfill
  \begin{minipage}[b]{0.48\textwidth}
    \centering
    \includegraphics[width=1.0\linewidth]{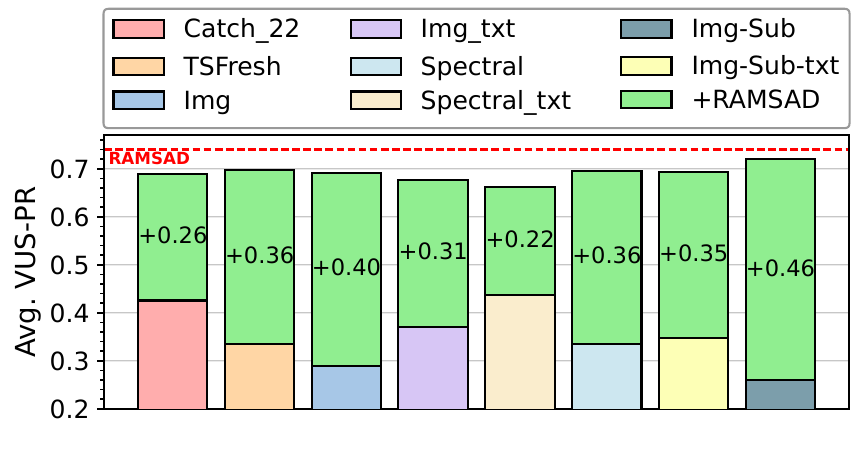}
    \caption{LLM-based model selection with and without RAG support (+RAMSAD). }
    \label{fig:llm_selection}
  \end{minipage}
\end{figure}

\textbf{Implication in Modern LLM-based Systems}.
{RAMSAD can also serve as the retrieval component (RAG) of LLM-based anomaly-detection pipelines.} 
To support this claim, Figure~\ref{fig:llm_selection} reports the average VUS-PR achieved when using a high-capacity LLM (mistral-large-3, 675B) as model selector, to which we provided different types of TS representations, including features (Catch 22~\cite{lubba2019catch22} and TSFresh~\cite{christ2018time}), images (full, spectral, and subsequence image), and LLM-generated summaries describing the latter (denoted as \textit{txt} in the figure). As we can see, the VUS-PR performance is rather low (between $0.28$ and $0.42$). 
Applying RAMSAD, we retrieved $5$ similar training examples and provided their detectors’ VUS-PR scores to the LLM. This drastically increased performances, with scores ranging from $0.66$ to $0.72$. While some of these results exceed the previous SOTA ($0.69$), they are still slightly lower than RAMSAD. 
RAMSAD supports post-hoc explainability, since retrieved neighbors and their detector-performance profiles can be given to ChatTS~\cite{xie2024chatts} to generate a rationale for the selected model.

{
\vspace{-0.2em}
\section{Conclusion}
}
We presented \textsc{RAMSAD}, a retrieval-based framework for model selection in time-series anomaly detection. 
By selecting detectors through similarity to previously observed series, \textsc{RAMSAD} avoids training a dedicated selector while remaining flexible and efficient{; its optional semantic alignment
is trained offline, independently of the selection task.}
Our key takeaways are as follows:

\textit{\textbf{Takeaway 1:} RAMSAD is the strongest overall method: it achieves the best single-model performance in ID and, when used to select small ensembles, reaches SOTA results in both ID and OOD while selecting substantially fewer detectors than full-pool ensemble methods. %
}

\textit{\textbf{Takeaway 2:} TSFM-based retrieval spaces are preferable for RAMSAD; among them, semantically aligned TSFM spaces deliver the strongest overall performance{, mainly through improved OOD
transfer, while unaligned TSFM spaces already outperform trained selectors.}}

These findings establish retrieval as a competitive foundation for anomaly-detector selection in heterogeneous time series.
We also observe (in preliminary experiments) that even when neighbours disagree, the averaged profile in RAMSAD still picks the right detector; we defer a detailed study to future work. 
Extending RAMSAD to multivariate series, temporal concept drift, and queries far from every covered domain are also interesting directions for future work (Appendix~\ref{app:limitations}).

\clearpage

\begin{ack}
Supported by EU Horizon project DataGEMS (101188416). This work was granted access to the HPC resources of IDRIS under the allocations 2025-A0191012641 and 2025-AD011016632 by GENCI. 
We thank Azedine Mani, Head of the IT team (UFR Math\'ematiques et Informatique, Universit\'e Paris Cit\'e), for his pivotal role in creating and making available computational resources essential to this work.
The authors declare no competing interests.

\end{ack}

\bibliographystyle{abbrvnat}
\bibliography{references}

\clearpage
\appendix
\appendixpage
\addappheadtotoc
\startcontents[sections]
\printcontents[sections]{l}{1}{\setcounter{tocdepth}{2}}

\clearpage
\section{Literature Review}
\label{sec:appendix_review}

Time-series anomaly detection (TSAD) has been studied extensively in recent years because of its importance in monitoring, diagnosis, and decision support across domains such as healthcare, manufacturing, finance, energy, and cybersecurity \cite{blazquezgarcia2022review,schmidl2022comprehensive}. 
A central challenge in TSAD is that anomalies may take very different forms, and the statistical properties of time series vary widely across datasets. 
As a result, the literature has produced a large variety of detectors, each relying on different assumptions about temporal structure, recurrence, noise, and anomaly manifestation. 
Recent surveys and benchmark studies consistently show that this diversity prevents the emergence of a single universally best detector across heterogeneous benchmark collections \cite{paparrizos2022tsb,schmidl2022comprehensive,liu2025tsb,sylligardosmsad2025}.

\textbf{Taxonomy of TSAD methods.}
A common organization of the TSAD literature distinguishes three broad families: \emph{distance-based}, \emph{density-based}, and \emph{prediction-based} methods \cite{blazquezgarcia2022review,schmidl2022comprehensive,sylligardosmsad2025}. 
Distance-based approaches identify anomalous observations or subsequences by measuring dissimilarity with respect to previously observed patterns. 
This family includes nearest-neighbor and discord-style methods, as well as matrix-profile-based approaches that search for subsequences with unusually large distances to their closest matches \cite{keogh2007compression,liu2009efficient,yeh2016matrix,linardi2020matrix}. 
These methods are often effective when anomalies correspond to rare subsequences that clearly depart from the dominant motifs of the series.

Density-based methods instead assess how isolated an observation or subsequence is in an induced representation space. 
Rather than relying only on raw temporal distances, they estimate local or global concentration patterns and flag sparse or weakly supported regions as anomalous. 
This category spans several subfamilies, including distribution-based approaches, graph-based methods, tree-based methods, and encoding-based methods \cite{antoni2019conditionIndicators,boniol2020series2graph,wenig2024series2graph++,liu2008isolation,ma2020isolation,Gao2020ensemble,sylligardosmsad2025}. 
Such methods are attractive when the notion of anomaly is better captured through structural relations or learned representations than through direct pointwise comparison.

Prediction-based methods model expected temporal behavior and use prediction error as the anomaly signal. 
They are commonly divided into \emph{forecasting-based} and \emph{reconstruction-based} approaches \cite{Wu_2022,su2019robust,sylligardosmsad2025}. 
Forecasting-based detectors predict future values from the past and treat large residuals as indicators of anomalous behavior, whereas reconstruction-based methods learn a compact representation of normal temporal patterns and detect anomalies via reconstruction error. 
These approaches are particularly appealing when normal dynamics are regular enough to be predicted or reconstructed reliably.

\textbf{Why detector performance varies across datasets.}
The lack of a universally dominant detector is now a well-established empirical finding in TSAD benchmarks \cite{paparrizos2022tsb,schmidl2022comprehensive,liu2025tsb}. 
A key reason is the heterogeneity of anomaly types. 
Following prior work, anomalies are often categorized as \emph{point}, \emph{contextual}, or \emph{collective} anomalies \cite{sylligardosmsad2025}. 
Point anomalies correspond to isolated values that strongly deviate from the bulk of the series, contextual anomalies are only abnormal relative to a local temporal context, and collective anomalies arise when a subsequence is abnormal as a whole even if individual points are not. 
Different detector families are naturally better suited to different anomaly types, which already makes detector performance highly dataset dependent.

A second source of variability comes from the structure of the time series itself. 
Even within the same anomaly category, datasets may differ in length, periodicity, stationarity, noise level, sampling characteristics, and the number and diversity of normal regimes \cite{schmidl2022comprehensive,sylligardosmsad2025}. 
For example, a detector that works well on time series with a single stable normal pattern may degrade substantially when applied to non-stationary data or to series containing multiple normalities. 
Similarly, methods that are effective for isolated rare events may behave differently when anomalies are frequent, repeated, or highly heterogeneous. 
These observations explain why benchmark studies repeatedly conclude that detector rankings are unstable across domains and that model choice should depend on the target series rather than on a single globally best detector \cite{paparrizos2022tsb,liu2025tsb}.

\textbf{Automatic detector selection.}
These benchmark findings have motivated a growing literature on \emph{automatic} approaches for choosing or combining detectors~\cite{liu2025tsb}, which is summarized in Figure~\ref{fig:autoML}. 
The first line of work is \emph{model selection}, where the objective is to choose the detector expected to perform best on a given time series. 
Many of these methods follow a meta-learning perspective: they use benchmark results together with representations of datasets or time series to learn how detector performance varies across problems. 
Classical algorithm-selection approaches such as \texttt{ARGOSMART} and \texttt{ISAC} rely on similarity in a meta-feature space, selecting either the best detector on the nearest training instance or the best detector inside a nearest cluster \cite{nikolic2013simple,kadioglu2010isac}. 
Other approaches formulate selection through collaborative filtering, regression, or classification, predicting detector identity or detector performance from dataset characteristics \cite{zhao2021automatic,xu2008satzilla,navarro2023meta,sylligardos2023choose,sylligardosmsad2025}.

A complementary group of methods avoids supervised meta-learning and instead evaluates candidate detectors directly on the target time series using internal criteria. 
These approaches include unsupervised score-evaluation techniques, clustering-based validation criteria, and detector-assessment schemes based on synthetic perturbations or rank aggregation \cite{goix2016evaluate,nguyen2016evaluation,chatterjee2022mospat,goswami2022unsupervised}. 
Their main appeal is that they do not require labeled historical performance data. 
However, they generally still require executing multiple detectors on the target series, which can substantially reduce their practical advantage when the candidate pool is large.

\textbf{Model ensembling and model generation.}
A second major line of work is \emph{model ensembling}, where multiple detectors are executed and their scores are aggregated to improve robustness \cite{aggarwal2015theoretical}. 
Simple averaging or max-based ensembles often provide a strong baseline because they hedge against detector-specific failures. 
More selective ensembling methods attempt to identify a subset of reliable detectors before aggregation \cite{rayana2016less,ma2023need}. 
At a larger systems level, \texttt{AutoTSAD} automates detector configuration, synthetic data generation, and final ensembling within a unified pipeline \cite{schmidl2024autotsad}. 
Although ensembles can be effective, their main limitation is computational: they typically require running many or all detectors before producing a final anomaly score.

A third direction is \emph{model generation}, where multiple detector outputs are used to build a new detector, usually through pseudo-labeling or boosting. 
Examples include \texttt{AutoOD-A} and \texttt{AutoOD-C}, which iteratively refine pseudo-labels and detector subsets, and \texttt{UADB}, which learns a booster from the outputs of several detectors \cite{cao2023autood,ye2023uadb}. 
These approaches can improve robustness and adaptivity, but they are also more complex, and like ensembling methods they often require evaluating a substantial fraction of the detector pool.

\textbf{Position of our work.}
Our work is most closely related to the \emph{model selection} line, but it departs from most recent TSAD selectors in an important way. 
Whereas much of the recent literature formulates detector selection as a trainable classification or regression problem \cite{sylligardos2023choose,sylligardosmsad2025}, we revisit a simpler non-parametric principle: similar time series should tend to prefer similar detectors. 
This intuition is related to earlier similarity-based algorithm selection methods such as \texttt{ARGOSMART} and \texttt{ISAC} \cite{nikolic2013simple,kadioglu2010isac}, but our framework is tailored to TSAD and is explicitly designed to operate directly on time-series similarity spaces, including raw-distance retrieval and embedding-based retrieval. 
In contrast to trainable selectors, retrieval-based model selection does not require fitting a separate predictive selector, can adapt naturally when the detector pool changes, and offers a simple way to leverage rich pretrained time-series representations.

Overall, the literature suggests two conclusions that motivate our approach. 
First, detector performance in TSAD is fundamentally heterogeneous, so automatic adaptation is necessary \cite{paparrizos2022tsb,liu2025tsb}. 
Second, while many existing automatic solutions are effective, they often rely either on heavy training pipelines or on the execution of many detectors at inference time \cite{sylligardosmsad2025,schmidl2024autotsad}. 
This leaves room for lightweight alternatives that exploit similarity structure in previously observed benchmark instances. 
Our proposed framework addresses this gap by casting detector selection as retrieval over a knowledge base of time series and detector-performance profiles.

\section{RAMSAD Method Details}
\label{app:method_details}

\subsection{Notation}
\label{app:notation}
Table~\ref{tab:symbols} summarizes the main notation used throughout the paper and appendix. 
We group the symbols into two parts. 
The first part introduces the general time-series anomaly-detection setting, including the collection of time series, candidate detectors, detector scores, labels, and performance measures. 
The second part defines the notation specific to RAMSAD, including the time-series and text encoders, segment and description embeddings, retrieval scores, the detector-performance matrix, and the semantic-alignment objective.

\begin{table}[hbt!]
\centering
\caption{Table of main symbols used in RAMSAD.}
\label{tab:symbols}
\resizebox{\columnwidth}{!}{%
\begin{tabular}{cl}
\toprule
\textbf{Symbol} & \textbf{Description} \\ 
\midrule
\multicolumn{2}{c}{\textit{General Notation}} \\
\midrule
$\mathfrak{D}$ & Collection of time series \\
$T_i, L_i$ & The $i$-th univariate time series and its length \\
$T_i^{(t,\ell)}$ & A subsequence of length $\ell$ extracted from $T_i$ starting at index $t$ \\
$\mathcal{B}$ & Finite pool of candidate anomaly detectors \\
$D_j$ & The $j$-th candidate anomaly detector \\
$S_{i,j}, Y_i$ & Anomaly score sequence produced by $D_j$ on $T_i$, and ground-truth labels \\
$Acc(\cdot)$ & A performance measure (e.g., AUC-PR, F1, VUS) \\
$\Phi, \Phi^\star$ & The model selector function, and the oracle choice \\
$M$ & The number of all detectors \\
$N$ & The number of detectors selected for an ensemble \\
\midrule
\multicolumn{2}{c}{\textit{RAMSAD Core Retrieval \& Alignment}} \\
\midrule
$f_{ts}, f_{text}$ & Time-series foundation model (encoder) and Text encoder \\
$z_{i, t}^{ts}, z_i^{desc}$ & Encoded representations of time-series segment and text description \\
$T_q, m$ & A query time series, and the number of segments extracted from it \\
$k$ & Number of retrieved neighbors \\
$\mathcal{A}(T_j)$ & Set of retrieved distances associated with segments from series $T_j$ \\
$base(T_j), freq(T_j)$ & Base similarity score and frequency score of candidate series $T_j$ \\
$score(T_j)$ & Final ranking score of candidate series $T_j$ \\
$\pi(\cdot)$ & Function mapping a segment to its originating time series \\
$top_k(T_q)$ & The final top-$k$ retrieved series for query $T_q$ \\
$\mathbf{V}$ & Precomputed performance matrix of detectors over the database \\
$\bar{V}_D(T_q)$ & Aggregated performance of detector $D$ over retrieved neighbors \\
$\widehat{D}(T_q)$ & The final selected detector for query $T_q$ \\
$\gamma$ & Similarity threshold for multi-positive matching \\
\(\tau\) & Temperature used in the contrastive alignment loss \\
$\mathbf{P}, \mathcal{L}$ & Binary multi-positive matching matrix, and symmetric alignment loss \\
$\mathcal{R}_i, \mathcal{Q}_j$ 
& Positive sets for time-series segment \(i\) and description \(j\) in the symmetric alignment loss \\

\bottomrule
\end{tabular}%
}
\end{table}

\subsection{Semantic Alignment Training}
\label{app:semantic_alignment}

\textbf{Trainable parameters.}
The time-series and text embeddings are precomputed using \texttt{Chronos2} and \texttt{all-mpnet-base-v2}, respectively. During alignment training, these backbone embeddings remain frozen, and only the projection heads are optimized. Each projection head is implemented as a lightweight multilayer perceptron of the architecture illustrated in Figure~\ref{fig:semantic_alignment_arc}
with hidden width equal to \(2\times\) the projection dimension. The outputs are \(\ell_2\)-normalized before computing similarities in the shared space.

\textbf{Optimization.}
Semantic alignment is trained with AdamW using learning rate \(10^{-5}\), weight decay \(10^{-4}\), and \(\beta=(0.9,0.999)\). Training runs for \(300\) epochs with gradient clipping at norm \(1.0\). We use a warmup-cosine learning-rate schedule, consisting of a linear warmup for the first \(10\) epochs followed by cosine decay to zero. The default batch size is \(8192\). The training split is capped at \(100{,}000\) samples, and examples are sampled using a stratified-uniform procedure so as to approximately balance the chosen stratification labels.

\textbf{Validation split and batching.}
The default train/validation split is \(0.8/0.2\), applied independently within each stratification group whenever possible. Stratification is based on the dataset-description column by default, with fallback to raw description text when needed.

\textbf{Loss configuration.}
The default alignment objective is the multi-positive loss described in the main paper. We use temperature \(0.1\) and define positive matches through description-embedding similarity, using a cosine-similarity threshold of \(0.95\). In addition, we mask same-series pairs so that positives must come from different time series within the batch. The framework also supports alternative loss functions such as contrastive, cosine, and distance-based objectives, though the reported experiments use the multi-positive setting.

{\textbf{Why multi-positive targets.}
Contrastive alignment is trained in batches, where each segment is paired with exactly one description, its positive. The model computes the similarity between each segment and every description in the batch, normalizes each row into a probability distribution, and minimizes the negative log-probability of the correct entry. Because each row must sum to one, raising the probability of the positive necessarily lowers that of every other entry ( i.e. every other description in the batch is implicitly treated as a negative and pushed away). This assumption breaks in our setting. For a segment from dataset \(A\), the description of dataset \(B\) is declared a negative even when \(A\) and \(B\) have near-synonymous descriptions (e.g., two distinct server-monitoring datasets). The objective then pushes the segment away from a semantically valid description, injecting label noise in proportion to the number of near-duplicate descriptions in the benchmark. We therefore treat pairs of descriptions whose frozen sentence-transformer embeddings have cosine similarity of at least \(\gamma=0.95\) as positives, which replaces the single correct answer per row by a multi-positive target matrix where a segment is no longer penalized for scoring highly against a description whose meaning matches its own. Note that this target matrix is computed from the frozen text embeddings only and it does not depend on the time-series embeddings or on the trainable projections, which only enter the predicted cross-modal similarities that the loss pushes toward the targets.}

{\textbf{Use of the trained heads at inference.}
Only the time-series projection head is used after training: knowledge-base segments and query segments are both encoded by the frozen TSFM, projected by the time-series head, and \(\ell_2\)-normalized, and retrieval compares them by cosine similarity. The text head and the text embeddings are neither stored in the index nor computed for a query. The alignment objective never uses detector identities or detector-performance values.}

{
\textbf{Symmetric Alignment Loss.}
The symmetric multi-positive objective can be understood as a batch-wise extension of standard contrastive learning. For each time-series segment \(i\), the denominator
\(
\sum_{j=1}^{B}\exp(s_{ij})
\)
normalizes the similarities to all descriptions in the batch, thereby defining a softmax distribution over candidate text matches. The numerator
\(
\sum_{j\in\mathcal{R}_i}\exp(s_{ij})
\)
collects the probability mass assigned to the subset of descriptions regarded as valid positives for that segment. The first summation term therefore encourages each segment embedding to assign high total probability to all compatible descriptions, rather than forcing the mass onto a single one-hot target.

The second term plays the same role in the reverse direction. For each description \(j\), the denominator
\(
\sum_{i=1}^{B}\exp(s_{ij})
\)
defines a softmax distribution over all time-series segments in the batch, while the numerator
\(
\sum_{i\in\mathcal{Q}_j}\exp(s_{ij})
\)
aggregates the mass assigned to the set of compatible segments. This enforces alignment both from segment to description and from description to segment, making the objective symmetric across modalities.

This multi-positive formulation is important in our setting because the supervision is inherently many-to-many. Different datasets may have highly similar descriptions, so several descriptions can be valid semantic matches for the same segment. Likewise, one description may be compatible with many segments from related time series. By summing over all positives in the numerator, the objective rewards the model whenever it concentrates probability mass on any semantically valid match, instead of penalizing it for not selecting one arbitrarily designated target. Since the softmax normalization couples positives and negatives, assigning more mass to valid matches necessarily reduces the mass assigned to incompatible pairs, so the loss remains contrastive.

Formally, letting \(s_{ij}=(z_i^{ts})^\top z_j^{desc}/\tau\) denote the temperature-scaled similarity between normalized projected embeddings, and letting \(\mathcal{R}_i\) and \(\mathcal{Q}_j\) denote the positive sets for segment \(i\) and description \(j\), respectively, the loss is
{
\[
\mathcal{L}
=
-\frac{1}{2B}
\left[
\sum_{i=1}^{B}
\log
\frac{\sum_{j \in \mathcal{R}_i}\exp(s_{ij})}
{\sum_{j=1}^{B}\exp(s_{ij})}
+
\sum_{j=1}^{B}
\log
\frac{\sum_{i \in \mathcal{Q}_j}\exp(s_{ij})}
{\sum_{i=1}^{B}\exp(s_{ij})}
\right].
\]
}
}

\textbf{Training loop and model selection.}
At each epoch, the model is trained on all mini-batches, evaluated on the validation set, and then updated by the scheduler. Validation performance is measured not only by loss but also by retrieval-style top-1 and top-5 accuracy in the shared space. More precisely, projected time-series and text embeddings are compared through matrix similarity, and retrieval is considered correct when the matched item belongs to the valid multi-positive set. Figure~\ref{fig:train_curves} reports the evolution of retrieval performance during training. The consistent increase in top-1 and top-5 accuracy across epochs indicates that the semantic alignment objective successfully improves cross-modal matching between time-series segments and dataset descriptions. The checkpoint with the lowest validation loss is stored.

The full semantic-alignment training procedure is given in Algorithm~\ref{algo:ramsad_align_full}. 
For each mini-batch, RAMSAD first encodes time-series segments and their associated textual descriptions using frozen pretrained encoders, and then maps both embeddings into a shared space through trainable projection heads. 
The algorithm constructs a multi-positive target matrix from in-batch description similarities, allowing semantically similar descriptions to act as valid positives rather than enforcing a single one-to-one match. 
After masking same-series pairs, the model optimizes the symmetric alignment loss described above, thereby learning a cross-modal space in which time-series segments and compatible descriptions are brought closer together.
\begin{figure}[tb]
    \centering
    \includegraphics[width=0.6\linewidth]{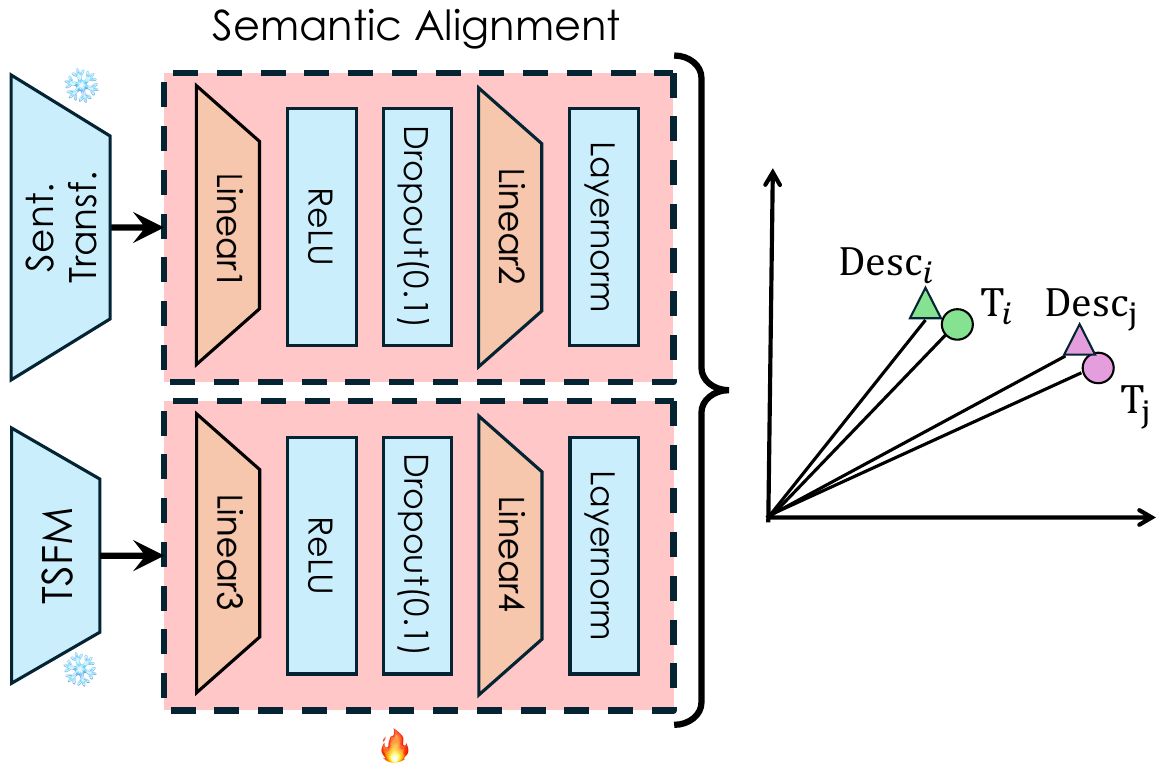}
    \caption{Semantic Alignment Architecture}
    \label{fig:semantic_alignment_arc}
\end{figure}

\algrenewcommand\algorithmicrequire{\textbf{Input:}}
\algrenewcommand\algorithmicensure{\textbf{Output:}}

\begin{algorithm}[t]
\caption{Semantic Alignment Training for RAMSAD}
\label{algo:ramsad_align_full}
\begin{algorithmic}[1]

\Require dataset \(\mathfrak{D}=\{T_1,\dots,T_{|\mathfrak{D}|}\}\), dataset descriptions \(\{c_i\}_{i=1}^{|\mathfrak{D}|}\), time-series encoder \(f_{\mathrm{ts}}\), text encoder \(f_{\mathrm{text}}\), projection heads \(g_{\theta^{\mathrm{ts}}}\) and \(g_{\theta^{\mathrm{desc}}}\), segment length \(\ell\), similarity threshold \(\gamma\), maximum number of epochs \(E_{\max}\), batch size \(b\), learning rate \(\eta\)
\Ensure learned projection parameters \(\theta^{\mathrm{ts}}, \theta^{\mathrm{desc}}\)

\State construct the set of segment--description pairs
\[
\mathcal{P}
=
\left\{
\left(T_i^{(t,\ell)}, c_i\right)
\;:\;
T_i \in \mathfrak{D},\;
T_i^{(t,\ell)} \text{ is a segment extracted from } T_i
\right\}
\]

\For{epoch \(=1,\dots,E_{\max}\)}

    \For{each mini-batch \(B=\{(T_{i_u}^{(t_u,\ell)},c_{i_u})\}_{u=1}^{b}\subset\mathcal{P}\)}

        \State encode time-series segments
        \[
        z^{\mathrm{ts}}_u
        =
        f_{\mathrm{ts}}\!\left(T_{i_u}^{(t_u,\ell)}\right),
        \qquad u=1,\dots,b
        \]

        \State encode descriptions using the frozen text encoder
        \[
        z^{\mathrm{desc}}_u
        =
        f_{\mathrm{text}}(c_{i_u}),
        \qquad u=1,\dots,b
        \]

        \State project both modalities into the shared space
        \[
        \tilde z^{\mathrm{ts}}_u = g_{\theta^{\mathrm{ts}}}\!\left(z^{\mathrm{ts}}_u\right),
        \qquad
        \tilde z^{\mathrm{desc}}_u = g_{\theta^{\mathrm{desc}}}\!\left(z^{\mathrm{desc}}_u\right)
        \]

        \State compute the in-batch description similarity matrix
        \[
        A_{uv}
        =
        \left\langle
        z^{\mathrm{desc}}_u, z^{\mathrm{desc}}_v
        \right\rangle
        \]

        \State build the multi-positive target matrix \(P \in \{0,1\}^{b\times b}\)
        \[
        P_{uv}
        =
        \mathbf{1}[A_{uv}\ge\gamma]
        \]

        \State mask pairs coming from the same source series

        \State compute the cross-modal similarity matrix
        \[
        S_{uv}
        =
        \left\langle
        \tilde z^{\mathrm{ts}}_u, \tilde z^{\mathrm{desc}}_v
        \right\rangle
        \]

        \State compute the symmetric cross-modal alignment loss \(\mathcal{L}(S,P)\)

        \State update \(\theta^{\mathrm{ts}}, \theta^{\mathrm{desc}}\) using \(\nabla \mathcal{L}\)

    \EndFor

\EndFor

\State \Return \(\theta^{\mathrm{ts}}, \theta^{\mathrm{desc}}\)

\end{algorithmic}
\end{algorithm}

\begin{figure}[tb]
  \centering

  \begin{subfigure}[t]{0.32\linewidth}
    \centering
    \includegraphics[trim={0 0 0 0.4cm},clip,width=\linewidth]{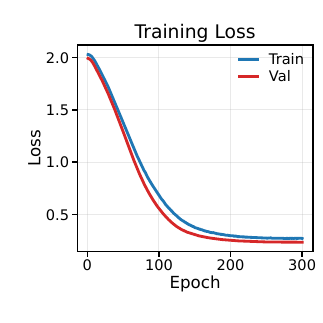}
    \caption{Train/Validation Loss}
  \end{subfigure}
  \hfill
  \begin{subfigure}[t]{0.32\linewidth}
    \centering
    \includegraphics[trim={0 0 0 0.4cm},clip,width=\linewidth]{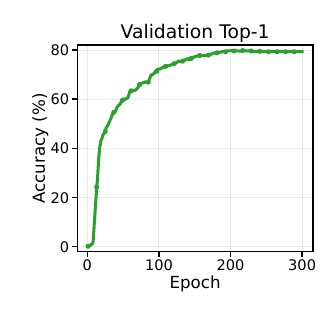}
    \caption{Top-1}
  \end{subfigure}
  \hfill
  \begin{subfigure}[t]{0.32\linewidth}
    \centering
    \includegraphics[trim={0 0 0 0.4cm},clip,width=\linewidth]{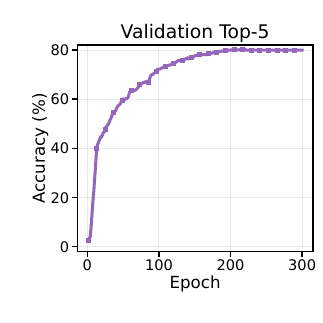}
    \caption{Top-5}
  \end{subfigure}

  \caption{As training evolves, the true series--description matches are found more effectively in their batch multi-positive similarity matrices.}
  \label{fig:train_curves}
\end{figure}

\subsection{Segment-to-Series Aggregation}
\label{app:segment_aggregation}

{Segment-level retrieval returns nearest neighbors for each query segment independently. 
However, detector selection in RAMSAD is performed at the level of full time series, since detector-performance profiles are stored for each training series rather than for individual segments. 
We therefore aggregate the segment-level neighbors into a ranked list of candidate series. 
The aggregation procedure counts how consistently a training series is retrieved across different query segments and combines this frequency information with the average retrieval distance of its matched segments. 
This favors training series that are both close to the query in representation space and repeatedly retrieved across multiple parts of the query series.

Algorithm~\ref{algo:ramsad_agg} describes this aggregation step. 
For each retrieved segment neighbor, we record its source time series and associated retrieval distance. 
For every candidate source series, RAMSAD computes a base distance score by averaging all distances assigned to that series, and a frequency score measuring the fraction of query segments for which the series appears among the retrieved neighbors. 
The final aggregation score rescales the average distance by the inverse of the frequency score, so that series retrieved more consistently across the query receive a better rank. 
The top-ranked series are then passed to the detector-selection step.}

\begin{algorithm}[t]
\caption{Segment-to-Series Aggregation}
\label{algo:ramsad_agg}
\begin{algorithmic}[1]

\Require query series \(T_q\); retrieved segment neighbors \(\{T_{j_{u,r}}^{(t_{u,r},\ell)}\}_{r=1}^{k}\) and distances \(\{d_{u,r}\}_{r=1}^{k}\) for each query segment \(T_q^{(t_u,\ell)}\), \(u=1,\dots,m\); small constant \(\epsilon>0\); retrieval size \(k\)
\Ensure ranked candidate series set \(\mathrm{top}_k(T_q)\)

\For{\textbf{all} query segments \(T_q^{(t_u,\ell)}\), \(u=1,\dots,m\)}
    \For{\textbf{all} retrieved neighbors \(T_{j_{u,r}}^{(t_{u,r},\ell)}\), \(r=1,\dots,k\)}
        \State append \(d_{u,r}\) to \(\mathcal{A}(T_{j_{u,r}})\)
    \EndFor
\EndFor

\For{\textbf{all} candidate series \(T_j\) such that \(\mathcal{A}(T_j)\neq\emptyset\)}
    \State compute
    \(
    \mathrm{base}(T_j)
    =
    \frac{1}{|\mathcal{A}(T_j)|}\sum_{d\in\mathcal{A}(T_j)} d
    \)
    \State compute
    \(
    \mathrm{freq}(T_j)
    =
    \frac{1}{m}
    \Bigl|
    \{\,u \in \{1,\dots,m\} : \exists r \in \{1,\dots,k\} \text{ s.t. } j_{u,r}=j \,\}
    \Bigr|
    \)
    \State compute
    \(
    \mathrm{score}(T_j)
    =
    \frac{1}{\max(\epsilon,\mathrm{freq}(T_j))}\cdot \mathrm{base}(T_j)
    \)
\EndFor

\State rank candidate series \(T_j \in \mathfrak{D}\) in ascending order of \(\mathrm{score}(T_j)\)
\State \Return the top-\(k\) ranked series, denoted by \(\mathrm{top}_k(T_q)\)

\end{algorithmic}
\end{algorithm}

\subsection{RAMSAD Model Selection}
\label{app:model_selection_algorithm}
{
After aggregating segment-level retrieval results into a ranked list of similar training series, RAMSAD uses the retrieved neighborhood to select a detector for the query series. 
The key assumption is that time series that are similar in the chosen representation space are likely to favor the same anomaly detector. 
Rather than training a separate classifier or regressor to predict the best detector, RAMSAD directly reuses the precomputed detector-performance profiles of the retrieved training series.

Algorithm~\ref{algo:ramsad_select} summarizes the full model-selection procedure. 
The query series is first segmented, encoded, and used to retrieve nearest-neighbor segments from the knowledge base. 
These segment-level neighbors are then aggregated into a ranked list of candidate training series using Algorithm~\ref{algo:ramsad_agg}. 
For each detector in the candidate pool, RAMSAD averages its stored performance values over the retrieved series and selects the detector with the highest aggregated score. 
This produces a non-parametric detector recommendation that can adapt to changes in the knowledge base or detector pool without retraining a selector model.}

\begin{algorithm}[H]
\small
\caption{RAMSAD Model Selection}
\label{algo:ramsad_select}
\begin{algorithmic}[1]
\algrenewcommand\algorithmicrequire{\textbf{Input:}}
\algrenewcommand\algorithmicensure{\textbf{Output:}}

\Require query series \(T_q\), retrieval index \(\mathcal{I}\), encoder \(f_{\mathrm{ts}}\), segment length \(\ell\), retrieval size \(k\), detector pool \(\mathcal{B}\), performance table \(\mathbf{V}\)
\Ensure selected detector \(\widehat{D}(T_q)\)

\State segment \(T_q\) into \(\{T_q^{(t_u,\ell)}\}_{u=1}^{m}\)
\State zero-pad the last segment if needed

\For{\textbf{all} segments \(T_q^{(t_u,\ell)}\)}
    \State encode \(T_q^{(t_u,\ell)}\) with \(f_{\mathrm{ts}}\)
    \State retrieve its top-\(k\) nearest database segments and their distances from \(\mathcal{I}\)
\EndFor

\State aggregate the retrieved segments into a ranked series list \(\mathrm{top}_k(T_q)\) using Algorithm~\ref{algo:ramsad_agg}

\For{\textbf{all} \(D \in \mathcal{B}\)}
    \State 
    {\(
    \bar{V}_D(T_q)
    =
    \frac{1}{|\mathrm{top}_k(T_q)|}\sum_{T \in \mathrm{top}_k(T_q)} \mathbf{V}(T,D)
    \)}
\EndFor

\State \(
\widehat{D}(T_q)
=
\arg\max_{D \in \mathcal{B}} \bar{V}_D(T_q)
\)

\State \Return \(\widehat{D}(T_q)\)

\end{algorithmic}
\end{algorithm}

\subsection{System Complexity and Scalability}
\label{app:system_complexity}

\textbf{Asymptotic retrieval cost.}
We analyze the computational cost of RAMSAD by comparing exact segment retrieval with an HNSW-based implementation. Let \(n\) denote the number of indexed knowledge-base segments, \(d\) the embedding dimensionality, \(b\) the number of query segments processed jointly, \(k_{seg}\) the number of retrieved segment neighbors per query segment, and \(k\) the number of final retrieved series \footnote{\(k<k_{seg}\) only if two or more query segments retrieve database segments that correspond to the same source series, otherwise \(k = k_{seg}\)}. For HNSW, we use \(deg\) for the maximum graph degree, \(\mathrm{ef}\) for the search candidate list size, \(\mathrm{ef}_c\) for the construction candidate list size, and \(L \approx \log_{deg} n\) for the number of graph layers. The dominant cost in RAMSAD comes from segment retrieval. With exact search, the index consists of the segment embedding matrix, requiring \(\mathcal{O}(nd)\) memory, together with the detector-performance matrix, which requires \(\mathcal{O}(|\mathfrak{D}|\,M)\) memory. Querying a batch of \(b\) segments requires computing all pairwise similarities against the indexed segments, followed by top-\(k_{seg}\) selection, for a cost dominated by \(\mathcal{O}(bnd)\). By contrast, HNSW adds an offline indexing stage and a small graph-memory overhead, but avoids scanning all \(n\) indexed segments at query time by performing graph traversal over a bounded candidate set.

\textbf{Cost of aggregation and recommendation.}
The remaining stages of RAMSAD are comparatively lightweight. After retrieval, the segment-to-series aggregation step processes at most \(bk_{seg}\) retrieved items, maps them to source series, and computes the score introduced in Section~\ref{sec:core_retrieval}. Its complexity is therefore linear in the size of the retrieval output and negligible relative to nearest-neighbor search. The final detector recommendation step is smaller still, since it only averages precomputed detector-performance values over the top-\(k\) retrieved series. Overall, the computational complexity of RAMSAD is governed by the retrieval backend rather than by aggregation or detector selection. The theoretical complexities are shown in Table~\ref{tab:complexity_theoretical}.

\setlength{\columnsep}{10pt} 
\begin{table}[t]
  \centering
  \small
  \caption{Asymptotic complexity of exact retrieval and HNSW within RAMSAD. Here, \(n\) is the number of indexed training segments, \(|\mathfrak{D}|\) the number of indexed training series, \(M\) the number of candidate detectors, \(d\) the embedding dimensionality, \(b\) the number of query segments processed jointly, and \(k_{seg}\) the segment-level retrieval size.}
  \label{tab:complexity_theoretical}
  \begin{tabular}{@{}llcc@{}}
    \toprule
    \textbf{Metric} & \textbf{} & \textbf{Exact} & \textbf{HNSW} \\
    \midrule
    \multirow{2}{*}{\textbf{Time}}
    & Build
    & ---
    & \(\mathcal{O}(nd(\mathrm{ef}_c + deg\,L))\) \\
    & Query
    & \(\mathcal{O}(bnd + bn + bk_{seg}\log k_{seg})\)
    & \(\mathcal{O}(b(\mathrm{ef_s}\,d + \mathrm{ef_s}\log\mathrm{ef_s}))\) \\
    \midrule
    \multirow{2}{*}{\textbf{Space}}
    & Persistent
    & \(\mathcal{O}(nd + |\mathfrak{D}|\,M)\)
    & \(\mathcal{O}(nd + n \ deg \ (L+1) + |\mathfrak{D}|\,M)\) \\
    & Query
    & \(\mathcal{O}(bd + bn + bk_{seg})\)
    & \(\mathcal{O}(bd + b\,\mathrm{ef} + bk_{seg})\) \\
    \bottomrule
  \end{tabular}
\end{table}

\section{Similarity Spaces: Performance and Runtime Analysis}
\label{app:similarity_spaces}

\subsection{Parameter Sensitivity with Respect to $k$}
\label{app:k_sensitivity}

Figure~\ref{fig:variation-k-all-models} provides the main visual summary of how performance varies with \(k\) across the different similarity spaces and TSFM backbones. 
Overall, the results show that RAMSAD is relatively stable with respect to the choice of \(k\), although the magnitude of this sensitivity differs across backbones and between ID and OOD settings. In the ID setting, performance varies only moderately across \(k\), with a maximum difference of roughly \(0.04\) across the tested configurations. 
This variability tends to decrease as the number of selected models increases, indicating that larger ensembles reduce the sensitivity to the retrieval size.
A common pattern is that intermediate values of \(k\) yield the strongest results, whereas very small or very large values are slightly less effective.

In the OOD setting, the variation across \(k\) is somewhat larger, reaching about \(0.07\) in the single-model case, but this difference also shrinks as the number of selected models increases. In contrast to the ID case, larger values of \(k\) tend to be consistently more favorable in OOD, suggesting that broader neighborhoods help stabilize model recommendation under domain shift. This behavior is particularly visible for the aligned representations, which generally exhibit clearer gains in OOD than in ID.

Among the evaluated backbones, \texttt{Chronos2}~\cite{ansari2025chronos2} and \texttt{Chronos2\_aligned} provide the strongest overall behavior, with alignment yielding the clearest gains in the OOD case. The other backbones---\texttt{Flowstate}~\cite{graf2025flowstate}, \texttt{Mantis}~\cite{feofanov2025mantis}, \texttt{Moirai}~\cite{woo2024moirai}, \texttt{Moment}~\cite{goswami2024moment}, \texttt{TimesFM}~\cite{das2023decoder}, \texttt{Tirex}~\cite{auer2025tirex}, and \texttt{Toto}~\cite{cohen2025toto}---follow the same broad pattern, but with different levels of sensitivity to \(k\) and different benefits from alignment. We therefore conclude that RAMSAD is robust to the choice of \(k\), while still benefiting from moderate-to-larger neighborhoods, especially in OOD settings, and that it consistently remains stronger than the best trainable selector baseline across this parameter range.

\begin{figure}[tb]
    \centering
    \includegraphics[width=0.7\linewidth]{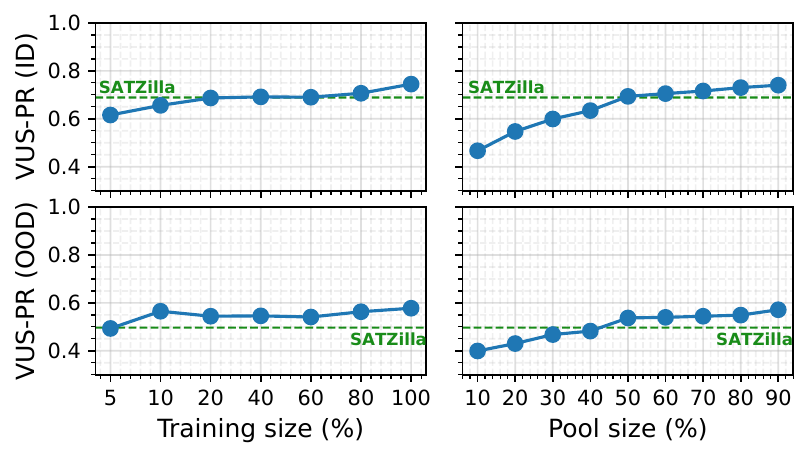}
    \caption{Average VUS-PR achieved by RAMSAD when randomly reducing the size of the training set (left) and detector pool (right) in ID (top) and OOD (bottom) scenarios.}
    \label{fig:reduced_sizes}
\end{figure}

\subsection{ID and OOD Performance Tables}
\label{app:id_ood_tables}
{
Tables~\ref{tab:similarity_id_k_merged} and~\ref{tab:similarity_ood_k_merged} report the complete numerical results corresponding to the parameter-sensitivity analysis Figure~\ref{fig:variation-k-all-models}.
For each similarity space, we report the mean, median, and standard deviation of the evaluation scores across different values of \(k\). 
The tables are organized so that raw and aligned variants of the same backbone are grouped together and separated from other backbones by double horizontal lines. 
This grouping makes it possible to compare, for each backbone, the effect of semantic alignment under the same retrieval setting.

Within each raw/aligned backbone group, bold values indicate the better result for a given statistic and value of \(k\). 
For the mean and median rows, larger values are better, whereas for the standard-deviation row, smaller values are better. 
Underlined values indicate the best overall value in the entire table for each statistic: the highest mean, the highest median, and the lowest standard deviation. }

\begin{table}
\caption{In-distribution. Bold per $k$ within each raw/aligned pair: higher mean and median, lower std. Underline marks the best value in the full table for each stat (max mean, max median, min std) across all methods and $k$.}
\label{tab:similarity_id_k_merged}
\centering
\scriptsize
\setlength{\tabcolsep}{4pt}
\renewcommand{\arraystretch}{0.95}
\begin{tabular}{llcccccccccc}
\toprule
\textbf{Method} & \textbf{Stat} & \textbf{k=1} & \textbf{k=2} & \textbf{k=3} & \textbf{k=4} & \textbf{k=5} & \textbf{k=6} & \textbf{k=7} & \textbf{k=8} & \textbf{k=9} & \textbf{k=10} \\
\midrule
\multirow{3}{*}{Chronos2} & Mean   & 0.7112 & 0.7117 & 0.7182 & 0.7329 & \textbf{0.7350} & \textbf{\underline{0.7463}} & \textbf{0.7422} & 0.7253 & \textbf{0.7216} & 0.7147 \\
 & Median & 0.8565 & 0.8552 & 0.8599 & 0.8854 & \textbf{0.8780} & \textbf{0.8925} & 0.8925 & \textbf{0.8891} & \textbf{0.8768} & 0.8565 \\
 & Std    & 0.3315 & 0.3329 & 0.3291 & 0.3198 & \textbf{0.3188} & \textbf{\underline{0.3105}} & \textbf{0.3177} & 0.3295 & \textbf{0.3274} & 0.3270 \\
\addlinespace[0.2pt]
\midrule
\addlinespace[0.2pt]
\multirow{3}{*}{Chronos2\_aligned} & Mean   & \textbf{0.7234} & \textbf{0.7233} & \textbf{0.7431} & \textbf{0.7457} & 0.7338 & 0.7384 & 0.7374 & \textbf{0.7260} & 0.7200 & \textbf{0.7285} \\
 & Median & \textbf{0.8854} & \textbf{0.8959} & \textbf{\underline{0.9024}} & \textbf{0.8925} & 0.8750 & 0.8854 & \textbf{0.8947} & 0.8867 & 0.8750 & \textbf{0.8867} \\
 & Std    & \textbf{0.3300} & \textbf{0.3324} & \textbf{0.3178} & \textbf{0.3126} & 0.3199 & 0.3173 & 0.3206 & \textbf{0.3278} & 0.3294 & \textbf{0.3269} \\
\addlinespace[1pt]
\midrule
\addlinespace[0.35pt]
\midrule
\multirow{3}{*}{Flowstate} & Mean   & 0.6263 & 0.6551 & 0.6724 & 0.6916 & 0.6911 & 0.6961 & 0.7032 & 0.6981 & 0.6983 & 0.6953 \\
 & Median & 0.7368 & 0.8043 & 0.8359 & 0.8334 & 0.8270 & 0.8410 & \textbf{0.8509} & 0.8453 & 0.8472 & 0.8410 \\
 & Std    & 0.3640 & 0.3659 & 0.3612 & 0.3411 & 0.3369 & 0.3404 & 0.3404 & 0.3417 & 0.3405 & 0.3382 \\
\addlinespace[0.2pt]
\midrule
\addlinespace[0.2pt]
\multirow{3}{*}{Flowstate\_aligned} & Mean   & \textbf{0.6408} & \textbf{0.6725} & \textbf{0.6888} & \textbf{0.7049} & \textbf{0.7099} & \textbf{0.7129} & \textbf{0.7118} & \textbf{0.7125} & \textbf{0.7093} & \textbf{0.7083} \\
 & Median & \textbf{0.7905} & \textbf{0.8174} & \textbf{0.8442} & \textbf{0.8445} & \textbf{0.8442} & \textbf{0.8496} & \textbf{0.8509} & \textbf{0.8580} & \textbf{0.8555} & \textbf{0.8552} \\
 & Std    & \textbf{0.3630} & \textbf{0.3536} & \textbf{0.3507} & \textbf{0.3325} & \textbf{0.3307} & \textbf{0.3327} & \textbf{0.3315} & \textbf{0.3362} & \textbf{0.3333} & \textbf{0.3321} \\
\addlinespace[1pt]
\midrule
\addlinespace[0.35pt]
\midrule
\multirow{3}{*}{Mantis} & Mean   & 0.6445 & \textbf{0.6718} & \textbf{0.6817} & \textbf{0.6896} & \textbf{0.6606} & \textbf{0.6806} & 0.6779 & \textbf{0.6759} & \textbf{0.6669} & \textbf{0.6700} \\
 & Median & 0.7982 & 0.8096 & \textbf{0.8335} & \textbf{0.8212} & \textbf{0.8017} & \textbf{0.8239} & \textbf{0.8270} & \textbf{0.8270} & \textbf{0.8169} & \textbf{0.8174} \\
 & Std    & 0.3677 & \textbf{0.3481} & \textbf{0.3456} & 0.3393 & 0.3539 & 0.3433 & 0.3464 & 0.3507 & 0.3539 & \textbf{0.3485} \\
\addlinespace[0.2pt]
\midrule
\addlinespace[0.2pt]
\multirow{3}{*}{Mantis\_aligned} & Mean   & \textbf{0.6601} & 0.6701 & 0.6732 & 0.6809 & 0.6592 & 0.6780 & \textbf{0.6824} & 0.6749 & 0.6655 & 0.6602 \\
 & Median & \textbf{0.8121} & \textbf{0.8136} & 0.8212 & 0.8092 & 0.8007 & 0.8169 & 0.8212 & \textbf{0.8270} & \textbf{0.8169} & 0.8096 \\
 & Std    & \textbf{0.3632} & 0.3498 & 0.3475 & \textbf{0.3371} & \textbf{0.3473} & \textbf{0.3378} & \textbf{0.3397} & \textbf{0.3489} & \textbf{0.3528} & 0.3521 \\
\addlinespace[1pt]
\midrule
\addlinespace[0.35pt]
\midrule
\multirow{3}{*}{Moirai} & Mean   & \textbf{0.6588} & \textbf{0.6845} & \textbf{0.6960} & \textbf{0.7116} & 0.6895 & \textbf{0.6867} & 0.6743 & 0.6697 & 0.6769 & \textbf{0.6834} \\
 & Median & \textbf{0.8092} & \textbf{0.8466} & \textbf{0.8552} & \textbf{0.8528} & \textbf{0.8466} & \textbf{0.8419} & 0.8419 & 0.8227 & 0.8320 & \textbf{0.8410} \\
 & Std    & \textbf{0.3616} & 0.3509 & 0.3481 & \textbf{0.3288} & 0.3458 & \textbf{0.3393} & 0.3499 & 0.3525 & 0.3460 & \textbf{0.3442} \\
\addlinespace[0.2pt]
\midrule
\addlinespace[0.2pt]
\multirow{3}{*}{Moirai\_aligned} & Mean   & 0.6486 & 0.6701 & 0.6925 & 0.6932 & \textbf{0.6945} & 0.6823 & \textbf{0.6961} & \textbf{0.6823} & \textbf{0.6827} & 0.6833 \\
 & Median & 0.7930 & 0.8043 & 0.8453 & 0.8419 & 0.8453 & 0.8418 & \textbf{0.8559} & \textbf{0.8438} & \textbf{0.8419} & 0.8320 \\
 & Std    & 0.3655 & \textbf{0.3469} & \textbf{0.3452} & 0.3419 & \textbf{0.3409} & 0.3441 & \textbf{0.3419} & \textbf{0.3478} & \textbf{0.3448} & 0.3457 \\
\addlinespace[1pt]
\midrule
\addlinespace[0.35pt]
\midrule
\multirow{3}{*}{Moment} & Mean   & \textbf{0.7157} & 0.7180 & 0.7084 & \textbf{0.7290} & \textbf{0.7171} & \textbf{0.7101} & \textbf{0.7103} & \textbf{0.7152} & \textbf{0.7087} & \textbf{0.7100} \\
 & Median & \textbf{0.8590} & 0.8584 & \textbf{0.8768} & \textbf{0.8765} & 0.8589 & \textbf{0.8582} & \textbf{0.8589} & \textbf{0.8637} & \textbf{0.8584} & \textbf{0.8580} \\
 & Std    & \textbf{0.3401} & 0.3345 & 0.3492 & 0.3309 & \textbf{0.3298} & \textbf{0.3329} & \textbf{0.3373} & \textbf{0.3344} & \textbf{0.3405} & \textbf{0.3381} \\
\addlinespace[0.2pt]
\midrule
\addlinespace[0.2pt]
\multirow{3}{*}{Moment\_aligned} & Mean   & 0.7064 & \textbf{0.7315} & \textbf{0.7160} & 0.7247 & 0.7109 & 0.7033 & 0.7042 & 0.7079 & 0.7050 & 0.6976 \\
 & Median & 0.8543 & \textbf{0.8637} & 0.8750 & 0.8590 & \textbf{0.8590} & 0.8543 & 0.8543 & 0.8565 & 0.8555 & 0.8509 \\
 & Std    & 0.3467 & \textbf{0.3276} & \textbf{0.3392} & \textbf{0.3282} & 0.3385 & 0.3378 & 0.3387 & 0.3412 & 0.3407 & 0.3437 \\
\addlinespace[1pt]
\midrule
\addlinespace[0.35pt]
\midrule
\multirow{3}{*}{TimesFM} & Mean   & 0.6809 & 0.6846 & 0.6904 & 0.6907 & 0.7084 & 0.7118 & 0.7143 & 0.6983 & 0.6855 & 0.6969 \\
 & Median & 0.8142 & 0.8142 & 0.8420 & 0.8295 & 0.8466 & \textbf{0.8496} & \textbf{0.8565} & 0.8453 & \textbf{0.8453} & \textbf{0.8466} \\
 & Std    & 0.3456 & 0.3424 & 0.3403 & 0.3380 & 0.3272 & 0.3232 & 0.3277 & 0.3354 & 0.3423 & 0.3355 \\
\addlinespace[0.2pt]
\midrule
\addlinespace[0.2pt]
\multirow{3}{*}{TimesFM\_aligned} & Mean   & \textbf{0.6961} & \textbf{0.7053} & \textbf{0.7075} & \textbf{0.7096} & \textbf{0.7118} & \textbf{0.7124} & \textbf{0.7146} & \textbf{0.7166} & \textbf{0.7083} & \textbf{0.7047} \\
 & Median & \textbf{0.8420} & \textbf{0.8322} & \textbf{0.8555} & \textbf{0.8431} & \textbf{0.8496} & 0.8466 & 0.8496 & \textbf{0.8496} & 0.8438 & 0.8463 \\
 & Std    & \textbf{0.3438} & \textbf{0.3293} & \textbf{0.3331} & \textbf{0.3253} & \textbf{0.3252} & \textbf{0.3223} & \textbf{0.3218} & \textbf{0.3208} & \textbf{0.3245} & \textbf{0.3287} \\
\addlinespace[1pt]
\midrule
\addlinespace[0.35pt]
\midrule
\multirow{3}{*}{Tirex} & Mean   & \textbf{0.6512} & 0.6588 & \textbf{0.6681} & 0.6753 & 0.6714 & \textbf{0.6652} & \textbf{0.6586} & 0.6573 & 0.6649 & 0.6701 \\
 & Median & \textbf{0.8095} & \textbf{0.8136} & \textbf{0.8295} & 0.8270 & 0.8212 & 0.8001 & \textbf{0.8001} & 0.8017 & 0.8249 & \textbf{0.8332} \\
 & Std    & \textbf{0.3652} & \textbf{0.3564} & \textbf{0.3563} & \textbf{0.3472} & \textbf{0.3433} & \textbf{0.3439} & \textbf{0.3508} & 0.3541 & 0.3497 & 0.3494 \\
\addlinespace[0.2pt]
\midrule
\addlinespace[0.2pt]
\multirow{3}{*}{Tirex\_aligned} & Mean   & 0.6337 & \textbf{0.6600} & 0.6453 & \textbf{0.6782} & \textbf{0.6812} & 0.6621 & 0.6518 & \textbf{0.6616} & \textbf{0.6770} & \textbf{0.6707} \\
 & Median & 0.7989 & 0.8092 & 0.8016 & \textbf{0.8295} & \textbf{0.8442} & \textbf{0.8286} & 0.7989 & \textbf{0.8174} & \textbf{0.8322} & 0.8249 \\
 & Std    & 0.3726 & 0.3576 & 0.3633 & 0.3479 & 0.3531 & 0.3566 & 0.3547 & \textbf{0.3513} & \textbf{0.3467} & \textbf{0.3481} \\
\addlinespace[1pt]
\midrule
\addlinespace[0.35pt]
\midrule
\multirow{3}{*}{Toto} & Mean   & \textbf{0.6786} & \textbf{0.7139} & \textbf{0.7087} & \textbf{0.7266} & 0.7159 & 0.7233 & 0.7130 & 0.7067 & 0.6973 & 0.6908 \\
 & Median & \textbf{0.8420} & \textbf{0.8555} & \textbf{0.8709} & \textbf{0.8637} & 0.8709 & \textbf{0.8719} & 0.8709 & 0.8584 & 0.8528 & 0.8528 \\
 & Std    & \textbf{0.3550} & 0.3362 & 0.3391 & 0.3225 & 0.3299 & \textbf{0.3190} & 0.3277 & 0.3351 & 0.3383 & 0.3433 \\
\addlinespace[0.2pt]
\midrule
\addlinespace[0.2pt]
\multirow{3}{*}{Toto\_aligned} & Mean   & 0.6719 & 0.7116 & 0.7050 & 0.7252 & \textbf{0.7282} & \textbf{0.7301} & \textbf{0.7271} & \textbf{0.7108} & \textbf{0.6996} & \textbf{0.7039} \\
 & Median & 0.8401 & 0.8466 & 0.8528 & 0.8599 & \textbf{0.8854} & \textbf{0.8719} & \textbf{0.8854} & \textbf{0.8599} & \textbf{0.8552} & \textbf{0.8552} \\
 & Std    & 0.3639 & \textbf{0.3295} & \textbf{0.3382} & \textbf{0.3221} & \textbf{0.3253} & 0.3194 & \textbf{0.3246} & \textbf{0.3324} & \textbf{0.3380} & \textbf{0.3351} \\
\addlinespace[1pt]
\midrule
\midrule
\addlinespace[0.2pt]
\multirow{3}{*}{Original (Euc.-seg.)} & Mean   & 0.6579 & 0.6694 & 0.6554 & 0.6565 & 0.6752 & 0.6532 & 0.6460 & 0.6123 & 0.6313 & 0.6148 \\
 & Median & 0.8092 & \textbf{0.8096} & 0.7903 & 0.7991 & 0.8282 & 0.7905 & 0.7733 & 0.7361 & 0.7502 & 0.7003 \\
 & Std    & 0.3598 & 0.3501 & 0.3517 & 0.3467 & \textbf{0.3423} & \textbf{0.3474} & 0.3472 & 0.3659 & 0.3536 & 0.3542 \\
\addlinespace[0.2pt]
\midrule
\addlinespace[0.2pt]
\multirow{3}{*}{Original (Euc. - full)} & Mean   & 0.5853 & 0.5913 & 0.5937 & 0.5775 & 0.5500 & 0.5582 & 0.5547 & 0.5528 & 0.5454 & 0.5173 \\
 & Median & 0.6936 & 0.7427 & 0.7148 & 0.6573 & 0.6055 & 0.6185 & 0.6064 & 0.6055 & 0.6161 & 0.5615 \\
 & Std    & 0.3808 & 0.3853 & 0.3823 & 0.3780 & 0.3831 & 0.3849 & 0.3829 & 0.3902 & 0.3855 & 0.3865 \\
\addlinespace[0.2pt]
\midrule
\addlinespace[0.2pt]
\multirow{3}{*}{Original (DTW-MESSI)} & Mean   & \textbf{0.6765} & \textbf{0.6880} & \textbf{0.6884} & \textbf{0.6861} & \textbf{0.6812} & \textbf{0.6775} & \textbf{0.6750} & \textbf{0.6643} & \textbf{0.6626} & \textbf{0.6506} \\
 & Median & \textbf{0.8333} & 0.8092 & \textbf{0.8333} & \textbf{0.8333} & \textbf{0.8373} & \textbf{0.8351} & \textbf{0.8286} & \textbf{0.8212} & \textbf{0.8286} & \textbf{0.8096} \\
 & Std    & \textbf{0.3554} & \textbf{0.3335} & \textbf{0.3350} & \textbf{0.3368} & 0.3440 & 0.3481 & \textbf{0.3435} & \textbf{0.3480} & \textbf{0.3532} & \textbf{0.3539} \\
\bottomrule
\end{tabular}
\end{table}

\begin{table}
\caption{Out-of-distribution. Bold per $k$ within each raw/aligned pair: higher mean and median, lower std. Underline marks the best value in the full table for each stat (max mean, max median, min std) across all methods and $k$.}
\label{tab:similarity_ood_k_merged}
\centering
\scriptsize
\setlength{\tabcolsep}{4pt}
\renewcommand{\arraystretch}{0.95}
\begin{tabular}{llcccccccccc}
\toprule
\textbf{Method} & \textbf{Stat} & \textbf{k=1} & \textbf{k=2} & \textbf{k=3} & \textbf{k=4} & \textbf{k=5} & \textbf{k=6} & \textbf{k=7} & \textbf{k=8} & \textbf{k=9} & \textbf{k=10} \\
\midrule
\multirow{3}{*}{Chronos2} & Mean   & 0.4819 & 0.4847 & 0.4900 & 0.5178 & 0.5427 & 0.5568 & 0.5721 & 0.5695 & 0.5917 & \textbf{0.5927} \\
 & Median & 0.4453 & 0.4477 & 0.4583 & 0.5837 & 0.6434 & 0.6435 & 0.6764 & 0.6764 & \textbf{\underline{0.7220}} & 0.6852 \\
 & Std    & 0.3844 & 0.3702 & 0.3834 & 0.3884 & 0.3809 & 0.3756 & 0.3683 & 0.3721 & 0.3704 & 0.3655 \\
\addlinespace[0.2pt]
\midrule
\addlinespace[0.2pt]
\multirow{3}{*}{Chronos2\_aligned} & Mean   & \textbf{0.5390} & \textbf{0.5162} & \textbf{0.5745} & \textbf{0.5768} & \textbf{0.5649} & \textbf{0.5822} & \textbf{0.5792} & \textbf{0.5899} & \textbf{\underline{0.5934}} & 0.5906 \\
 & Median & \textbf{0.6073} & \textbf{0.5291} & \textbf{0.6794} & \textbf{0.6578} & \textbf{0.6578} & \textbf{\underline{0.7220}} & \textbf{0.6852} & \textbf{0.7124} & \textbf{\underline{0.7220}} & \textbf{0.6932} \\
 & Std    & 0.3819 & 0.3771 & 0.3817 & 0.3784 & 0.3844 & 0.3805 & 0.3803 & 0.3764 & 0.3750 & 0.3757 \\
\addlinespace[1pt]
\midrule
\addlinespace[0.35pt]
\midrule
\multirow{3}{*}{Flowstate} & Mean   & 0.4932 & 0.4876 & \textbf{0.5237} & \textbf{0.5088} & \textbf{0.5311} & \textbf{0.5260} & \textbf{0.5380} & 0.5160 & 0.5217 & \textbf{0.5383} \\
 & Median & 0.5228 & 0.5307 & \textbf{0.6032} & \textbf{0.5707} & \textbf{0.6064} & \textbf{0.6185} & \textbf{0.6435} & 0.5850 & 0.6000 & \textbf{0.6360} \\
 & Std    & 0.3872 & 0.3860 & 0.3918 & 0.3874 & 0.3852 & 0.3861 & 0.3879 & 0.3846 & 0.3865 & 0.3835 \\
\addlinespace[0.2pt]
\midrule
\addlinespace[0.2pt]
\multirow{3}{*}{Flowstate\_aligned} & Mean   & \textbf{0.4962} & \textbf{0.4975} & 0.5058 & 0.4961 & 0.5157 & 0.5138 & 0.5335 & \textbf{0.5360} & \textbf{0.5386} & 0.5374 \\
 & Median & \textbf{0.5392} & \textbf{0.5409} & 0.5409 & 0.5300 & 0.5956 & 0.5850 & 0.6236 & \textbf{0.6366} & \textbf{0.6452} & 0.6236 \\
 & Std    & 0.3856 & 0.3922 & 0.3932 & 0.3873 & 0.3865 & 0.3914 & 0.3930 & 0.3922 & 0.3892 & 0.3848 \\
\addlinespace[1pt]
\midrule
\addlinespace[0.35pt]
\midrule
\multirow{3}{*}{Mantis} & Mean   & 0.3751 & 0.4001 & \textbf{0.4208} & \textbf{0.4058} & 0.3880 & \textbf{0.3901} & 0.3960 & \textbf{0.4131} & \textbf{0.4046} & 0.3846 \\
 & Median & 0.2032 & 0.2948 & \textbf{0.3559} & \textbf{0.3245} & \textbf{0.2889} & \textbf{0.2889} & 0.2962 & \textbf{0.3308} & \textbf{0.3005} & 0.2671 \\
 & Std    & 0.3701 & 0.3709 & 0.3652 & 0.3610 & 0.3544 & 0.3514 & 0.3537 & 0.3579 & 0.3569 & 0.3542 \\
\addlinespace[0.2pt]
\midrule
\addlinespace[0.2pt]
\multirow{3}{*}{Mantis\_aligned} & Mean   & \textbf{0.3779} & \textbf{0.4171} & 0.3990 & 0.3940 & \textbf{0.3886} & 0.3855 & \textbf{0.4111} & 0.3903 & 0.3985 & \textbf{0.4041} \\
 & Median & \textbf{0.2808} & \textbf{0.3333} & 0.2953 & 0.3053 & 0.2779 & 0.2835 & \textbf{0.3360} & 0.2892 & 0.2936 & \textbf{0.3005} \\
 & Std    & 0.3612 & 0.3683 & 0.3610 & 0.3571 & 0.3554 & 0.3513 & 0.3537 & 0.3499 & 0.3566 & 0.3561 \\
\addlinespace[1pt]
\midrule
\addlinespace[0.35pt]
\midrule
\multirow{3}{*}{Moirai} & Mean   & 0.4676 & 0.4923 & 0.5070 & 0.5136 & \textbf{0.5257} & 0.5363 & \textbf{0.5383} & 0.5256 & 0.4938 & 0.4956 \\
 & Median & 0.4091 & 0.4302 & 0.5092 & 0.5466 & \textbf{0.6158} & 0.6258 & \textbf{0.6258} & 0.5956 & 0.4765 & 0.5092 \\
 & Std    & 0.3798 & 0.3800 & 0.3843 & 0.3900 & 0.3895 & 0.3880 & 0.3872 & 0.3935 & 0.3973 & 0.3951 \\
\addlinespace[0.2pt]
\midrule
\addlinespace[0.2pt]
\multirow{3}{*}{Moirai\_aligned} & Mean   & \textbf{0.4952} & \textbf{0.4975} & \textbf{0.5132} & \textbf{0.5161} & 0.5176 & \textbf{0.5377} & 0.5300 & \textbf{0.5305} & \textbf{0.5060} & \textbf{0.5198} \\
 & Median & \textbf{0.4477} & \textbf{0.4453} & \textbf{0.5334} & \textbf{0.5545} & 0.5697 & \textbf{0.6353} & 0.5956 & \textbf{0.6158} & \textbf{0.5412} & \textbf{0.5778} \\
 & Std    & 0.3861 & 0.3788 & 0.3823 & 0.3928 & 0.3901 & 0.3888 & 0.3884 & 0.3885 & 0.3911 & 0.3859 \\
\addlinespace[1pt]
\midrule
\addlinespace[0.35pt]
\midrule
\multirow{3}{*}{Moment} & Mean   & \textbf{0.4711} & \textbf{0.4578} & 0.4434 & 0.4504 & 0.4419 & 0.4443 & 0.4529 & 0.4537 & 0.4592 & 0.4453 \\
 & Median & \textbf{0.4399} & \textbf{0.4358} & 0.4042 & \textbf{0.4091} & 0.3687 & 0.3757 & 0.4001 & \textbf{0.4042} & \textbf{0.4215} & 0.3761 \\
 & Std    & 0.3760 & 0.3653 & 0.3635 & 0.3551 & 0.3549 & 0.3491 & 0.3539 & 0.3516 & 0.3509 & 0.3489 \\
\addlinespace[0.2pt]
\midrule
\addlinespace[0.2pt]
\multirow{3}{*}{Moment\_aligned} & Mean   & 0.4581 & 0.4575 & \textbf{0.4678} & \textbf{0.4523} & \textbf{0.4520} & \textbf{0.4559} & \textbf{0.4581} & \textbf{0.4611} & \textbf{0.4602} & \textbf{0.4582} \\
 & Median & 0.4023 & 0.3946 & \textbf{0.4073} & 0.3983 & \textbf{0.4007} & \textbf{0.4023} & \textbf{0.4027} & 0.4027 & 0.4027 & \textbf{0.4027} \\
 & Std    & 0.3752 & 0.3725 & 0.3617 & 0.3525 & 0.3527 & 0.3515 & 0.3511 & 0.3513 & 0.3531 & 0.3498 \\
\addlinespace[1pt]
\midrule
\addlinespace[0.35pt]
\midrule
\multirow{3}{*}{TimesFM} & Mean   & \textbf{0.4772} & 0.4541 & 0.4738 & 0.4660 & 0.4731 & 0.4652 & 0.4693 & 0.4654 & 0.4830 & 0.4916 \\
 & Median & 0.3757 & 0.3687 & 0.3977 & 0.3755 & 0.3946 & 0.3777 & 0.3910 & 0.3910 & 0.4341 & 0.4219 \\
 & Std    & 0.3737 & 0.3697 & 0.3634 & 0.3714 & 0.3602 & 0.3555 & 0.3551 & 0.3592 & 0.3614 & 0.3609 \\
\addlinespace[0.2pt]
\midrule
\addlinespace[0.2pt]
\multirow{3}{*}{TimesFM\_aligned} & Mean   & 0.4525 & \textbf{0.4636} & \textbf{0.4876} & \textbf{0.4898} & \textbf{0.4922} & \textbf{0.4879} & \textbf{0.4854} & \textbf{0.4883} & \textbf{0.4994} & \textbf{0.4965} \\
 & Median & \textbf{0.3761} & \textbf{0.4041} & \textbf{0.4437} & \textbf{0.5092} & \textbf{0.5092} & \textbf{0.4903} & \textbf{0.5063} & \textbf{0.5256} & \textbf{0.5307} & \textbf{0.5269} \\
 & Std    & 0.3702 & 0.3556 & 0.3676 & 0.3662 & 0.3608 & 0.3691 & 0.3631 & 0.3656 & 0.3665 & 0.3723 \\
\addlinespace[1pt]
\midrule
\addlinespace[0.35pt]
\midrule
\multirow{3}{*}{Tirex} & Mean   & \textbf{0.4985} & \textbf{0.4961} & 0.4705 & \textbf{0.4874} & \textbf{0.4941} & \textbf{0.4960} & \textbf{0.4871} & \textbf{0.4921} & \textbf{0.4966} & \textbf{0.5068} \\
 & Median & \textbf{0.4852} & \textbf{0.4904} & 0.4023 & 0.4558 & 0.4551 & \textbf{0.5198} & 0.4154 & \textbf{0.4659} & \textbf{0.4854} & \textbf{0.5258} \\
 & Std    & 0.3731 & 0.3699 & 0.3678 & 0.3705 & 0.3728 & 0.3719 & 0.3740 & 0.3761 & 0.3702 & 0.3671 \\
\addlinespace[0.2pt]
\midrule
\addlinespace[0.2pt]
\multirow{3}{*}{Tirex\_aligned} & Mean   & 0.4698 & 0.4705 & \textbf{0.4794} & 0.4784 & 0.4812 & 0.4906 & 0.4815 & 0.4751 & 0.4790 & 0.4826 \\
 & Median & 0.4011 & 0.3983 & \textbf{0.4109} & \textbf{0.4568} & \textbf{0.4568} & 0.4663 & \textbf{0.4188} & 0.4109 & 0.4390 & 0.4185 \\
 & Std    & 0.3776 & 0.3813 & 0.3805 & 0.3835 & 0.3828 & 0.3814 & 0.3863 & 0.3864 & 0.3836 & 0.3835 \\
\addlinespace[1pt]
\midrule
\addlinespace[0.35pt]
\midrule
\multirow{3}{*}{Toto} & Mean   & \textbf{0.4447} & 0.4455 & \textbf{0.4459} & 0.4360 & 0.4196 & \textbf{0.4456} & 0.4495 & \textbf{0.4527} & \textbf{0.4578} & \textbf{0.4545} \\
 & Median & 0.3376 & 0.3826 & \textbf{0.4023} & 0.3477 & 0.3103 & \textbf{0.3777} & \textbf{0.3790} & \textbf{0.3826} & \textbf{0.4207} & \textbf{0.3983} \\
 & Std    & 0.3866 & 0.3768 & 0.3743 & 0.3768 & 0.3725 & 0.3702 & 0.3742 & 0.3747 & 0.3713 & 0.3778 \\
\addlinespace[0.2pt]
\midrule
\addlinespace[0.2pt]
\multirow{3}{*}{Toto\_aligned} & Mean   & 0.4383 & \textbf{0.4542} & 0.4331 & \textbf{0.4464} & \textbf{0.4647} & 0.4332 & \textbf{0.4498} & 0.4429 & 0.4548 & 0.4517 \\
 & Median & \textbf{0.3440} & \textbf{0.4104} & 0.3512 & \textbf{0.3738} & \textbf{0.4041} & 0.3168 & 0.3533 & 0.3402 & 0.3738 & 0.3738 \\
 & Std    & 0.3791 & 0.3759 & 0.3772 & 0.3710 & 0.3792 & 0.3821 & 0.3841 & 0.3769 & 0.3821 & 0.3848 \\
\addlinespace[1pt]
\midrule
\midrule
\addlinespace[0.2pt]
\multirow{3}{*}{Original (Euc.-seg.)} & Mean   & 0.3862 & 0.3952 & 0.3811 & 0.3913 & 0.3960 & 0.4098 & 0.4141 & 0.4133 & 0.4232 & 0.4218 \\
 & Median & 0.2744 & 0.2953 & 0.2840 & 0.2943 & 0.3168 & 0.3303 & \textbf{0.3333} & 0.3400 & \textbf{0.3694} & 0.3482 \\
 & Std    & \textbf{0.3637} & \textbf{0.3523} & \textbf{0.3473} & \textbf{\underline{0.3452}} & \textbf{0.3474} & \textbf{0.3511} & \textbf{0.3534} & \textbf{0.3538} & \textbf{0.3602} & 0.3610 \\
\addlinespace[0.2pt]
\midrule
\addlinespace[0.2pt]
\multirow{3}{*}{Original (Euc. - full)} & Mean   & 0.3707 & 0.4392 & 0.4178 & 0.4151 & 0.3990 & \textbf{0.4256} & 0.4140 & 0.3971 & 0.4039 & 0.3941 \\
 & Median & 0.2046 & 0.3510 & 0.3006 & 0.2983 & 0.2892 & \textbf{0.3561} & \textbf{0.3333} & 0.3115 & 0.3245 & 0.3035 \\
 & Std    & 0.3645 & 0.3789 & 0.3705 & 0.3739 & 0.3625 & 0.3693 & 0.3629 & 0.3567 & 0.3628 & \textbf{0.3594} \\
\addlinespace[0.2pt]
\midrule
\addlinespace[0.2pt]
\multirow{3}{*}{Original (DTW-MESSI)} & Mean   & \textbf{0.4229} & \textbf{0.4396} & \textbf{0.4276} & \textbf{0.4304} & \textbf{0.4220} & 0.4197 & \textbf{0.4215} & \textbf{0.4295} & \textbf{0.4365} & \textbf{0.4354} \\
 & Median & \textbf{0.3482} & \textbf{0.3579} & \textbf{0.3606} & \textbf{0.4023} & \textbf{0.3385} & 0.3128 & 0.3192 & \textbf{0.3482} & 0.3588 & \textbf{0.3651} \\
 & Std    & 0.3696 & 0.3776 & 0.3668 & 0.3670 & 0.3702 & 0.3780 & 0.3762 & 0.3769 & 0.3782 & 0.3786 \\
\bottomrule
\end{tabular}
\end{table}

\subsection{Retrieval Runtime Across TSFM Similarity Spaces}
\label{app:similarity_runtime}

Tables~\ref{tab:tsfm_runtime_simplified} and~\ref{tab:raw_measures_runtime_simple} compare the wall-clock time of the RAMSAD retrieval pipeline across the similarity spaces considered in this work. The goal here is to isolate the computational cost induced by the choice of representation. All methods are therefore evaluated on a common query--database pool, and the reported times should be interpreted as single-run retrieval costs.

Table~\ref{tab:tsfm_runtime_simplified} reports the runtime decomposition for TSFM-based and aligned TSFM-based similarity spaces. A substantial part of the variation across backbones is explained by the dimensionality of the embeddings they produce, since both loading and nearest-neighbor search scale with the size of the representation. In particular, backbones with smaller native embedding dimensions tend to be faster: \texttt{Tirex} is the fastest (\(99.9\) ms), followed by \texttt{Mantis} (\(110.8\) ms), which uses the lowest-dimensional representation among the tested TSFMs, whereas \texttt{TimesFM}, which produces the largest embeddings, is the slowest (\(144.9\) ms). The aligned variants introduce an additional projection step to a shared 768-dimensional space, which costs between \(17\) and \(25\) ms depending on the backbone. Despite this overhead, all aligned variants remain well below one second, showing that semantic alignment improves the representation space at only moderate additional cost. In particular, \texttt{Chronos-2\_aligned}, which provides the strongest retrieval quality in our experiments, requires \(129.1\) ms and therefore remains practical.

\begin{table}[t]
\caption{Retrieval-pipeline wall-clock time (ms) across TSFM-based similarity spaces. All methods are evaluated on the same query and database pool (\(N_{\text{test}}=1383\), \(N_{\text{train}}=3264\), \(N_{\text{pairs}}=4{,}514{,}112\)). Aligned variants apply the learned 768-dimensional semantic projection before retrieval.}
\label{tab:tsfm_runtime_simplified}
\centering
\small
\setlength{\tabcolsep}{4pt}
\begin{tabular}{lrrrrr}
\toprule
\textbf{Method} & \textbf{Load} & \textbf{Proj.} & \textbf{Search} & \textbf{Agg.} & \textbf{Total} \\
\midrule
Mantis          & 10.0 & --   & 44.6 & 56.2 & 110.8 \\
Tirex           & 12.3 & --   & 47.4 & 40.2 & 99.9 \\
Flowstate       & 14.1 & --   & 44.8 & 60.8 & 119.7 \\
Moirai          & 14.6 & --   & 56.0 & 53.0 & 123.6 \\
Toto            & 14.8 & --   & 61.0 & 50.7 & 126.5 \\
Moment          & 17.7 & --   & 53.9 & 54.1 & 125.7 \\
Chronos-2       & 16.4 & --   & 69.0 & 52.6 & 138.0 \\
TimesFM         & 25.2 & --   & 64.3 & 55.4 & 144.9 \\
\midrule
Mantis\_aligned     & 10.0 & 18.7 & 44.6 & 56.2 & 129.5 \\
Tirex\_aligned      & 12.3 & 18.1 & 50.3 & 57.6 & 138.3 \\
Flowstate\_aligned  & 14.1 & 17.0 & 51.5 & 56.4 & 139.0 \\
Moirai\_aligned     & 14.6 & 21.3 & 49.5 & 56.2 & 141.6 \\
Toto\_aligned       & 16.8 & 20.5 & 51.6 & 55.9 & 144.8 \\
Moment\_aligned     & 17.7 & 23.2 & 56.8 & 54.0 & 151.7 \\
Chronos-2\_aligned  & 14.8 & 25.3 & 48.9 & 40.1 & 129.1 \\
TimesFM\_aligned    & 25.6 & 23.2 & 49.8 & 55.9 & 154.5 \\
\bottomrule
\end{tabular}
\end{table}

\subsection{Retrieval Runtime of Original Similarity Space}
\label{app:raw_distance_runtime}

Table~\ref{tab:raw_measures_runtime_simple} places these results in the context of the original similarity spaces. Segment-level Euclidean retrieval with aggregation requires \(549.2\) ms, while full-series Euclidean retrieval requires \(1.20\) s. By contrast, DTW retrieval via MESSI~\cite{peng2021fast, del2026daisy}\footnote{\url{https://github.com/MChatzakis/DaiSy}} using a warping window of \(10\%\) is several orders of magnitude slower, requiring \(11\) h \(43\) m on the same pool. This confirms that, although DTW can be competitive in accuracy among original-space distances, its computational cost makes it impractical as a retrieval backend for RAMSAD.

Overall, these results reinforce the main conclusion of this section: embedding-based similarity spaces, and especially aligned TSFM spaces, provide the best practical accuracy--efficiency trade-off for RAMSAD.

\begin{table}[t]
\caption{Retrieval-pipeline wall-clock time of original similarity spaces on the same query and database pool.}
\label{tab:raw_measures_runtime_simple}
\centering
\small
\setlength{\tabcolsep}{5pt}
\begin{tabular}{lr}
\toprule
\textbf{Method} & \textbf{Total time} \\
\midrule
Segment Euclidean + aggregation & 549.2\,ms \\
Full-series Euclidean (truncated) & 1.20\,s \\
DTW via MESSI (\(w=10\%\)) & 11\,h\,43\,m \\
\bottomrule
\end{tabular}
\end{table}

\begin{table}[t]
\caption{OOD retrieval wall-clock time per domain for the three raw-distance similarity spaces.}
\label{tab:raw_measures_ood_domain_simple}
\centering
\small
\setlength{\tabcolsep}{6pt}
\renewcommand{\arraystretch}{1.12}
\begin{tabular}{lrrr}
\toprule
\textbf{Domain} & \textbf{Seg. Eucl.} & \textbf{Full-series Eucl.} & \textbf{DTW (MESSI)} \\
\midrule
Environment    & 189.1\,ms & 106.0\,ms & 30\,m\,46\,s \\
Facility       & 575.0\,ms & 1.19\,s   & 14\,h\,42\,m \\
Finance        & 189.2\,ms & 126.3\,ms & 37\,m\,18\,s \\
Human Activity & 292.0\,ms & 557.0\,ms & 5\,h\,29\,m \\
Medical        & 616.6\,ms & 1.24\,s   & 12\,h\,29\,m \\
Sensor         & 211.9\,ms & 235.2\,ms & 2\,h\,20\,m \\
Synthetic      & 218.2\,ms & 467.1\,ms & 2\,h\,12\,m \\
Traffic        & 184.0\,ms & 24.7\,ms  & 1\,m\,56\,s \\
Web Service    & 425.3\,ms & 1.05\,s   & 6\,h\,08\,m \\
\midrule
\textbf{Total} & \textbf{2.90\,s} & \textbf{5.00\,s} & \textbf{44\,h\,32\,m} \\
\bottomrule
\end{tabular}
\end{table}

\begin{figure}[tb]
  \centering

  \begin{subfigure}[t]{0.4\textwidth}
    \centering
    \includegraphics[width=\linewidth]{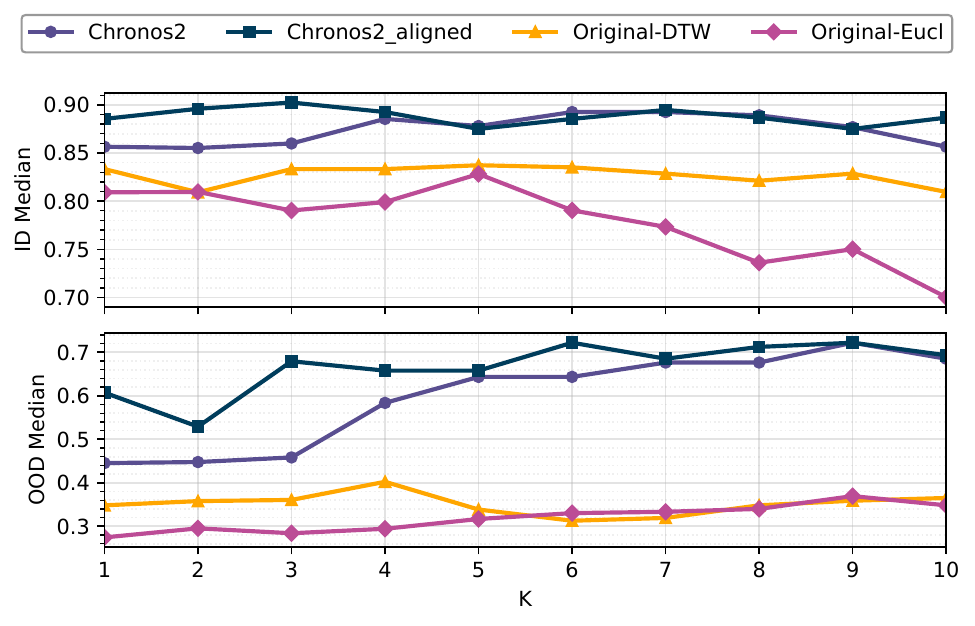}
    \vspace{-0.7cm}
    \caption{Chronos2}
  \end{subfigure}
  \hfill
  \begin{subfigure}[t]{0.4\textwidth}
    \centering
    \includegraphics[width=\linewidth]{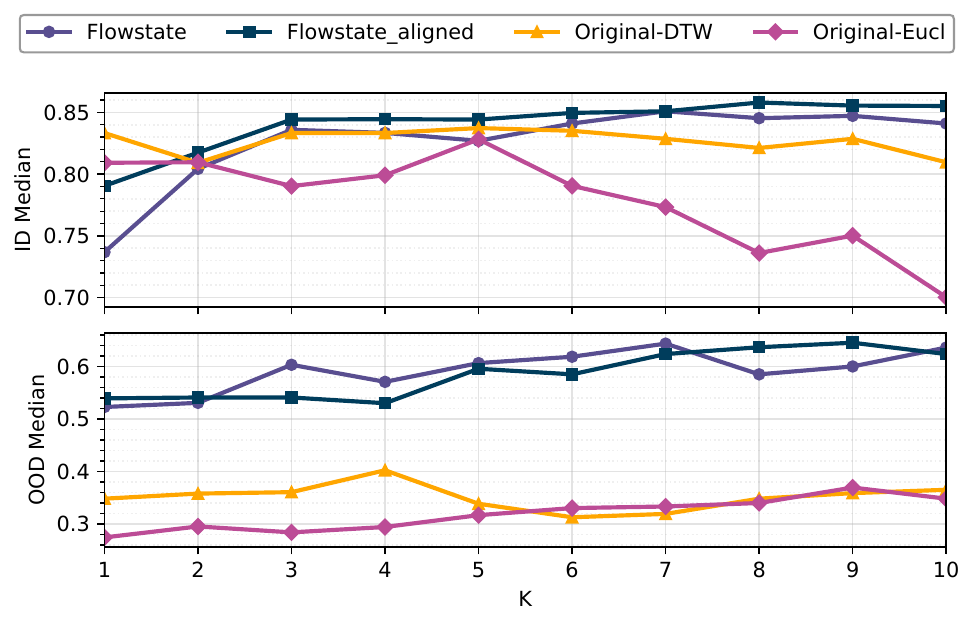}
    \vspace{-0.7cm}
    \caption{Flowstate}
  \end{subfigure}

  \vspace{0.2em}

  \begin{subfigure}[t]{0.4\textwidth}
    \centering
    \includegraphics[width=\linewidth]{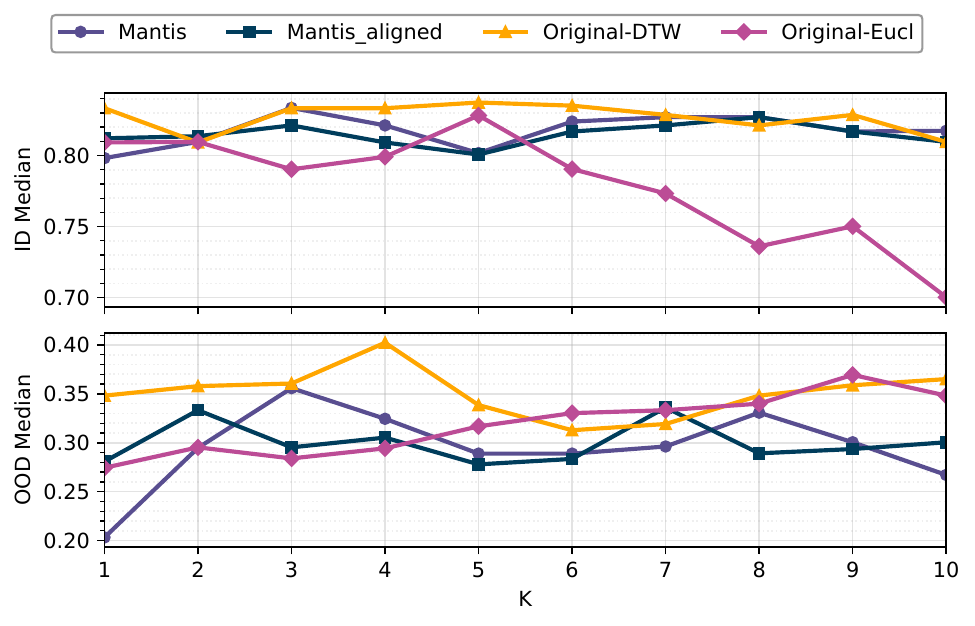}
    \vspace{-0.7cm}
    \caption{Mantis}
  \end{subfigure}
  \hfill
  \begin{subfigure}[t]{0.4\textwidth}
    \centering
    \includegraphics[width=\linewidth]{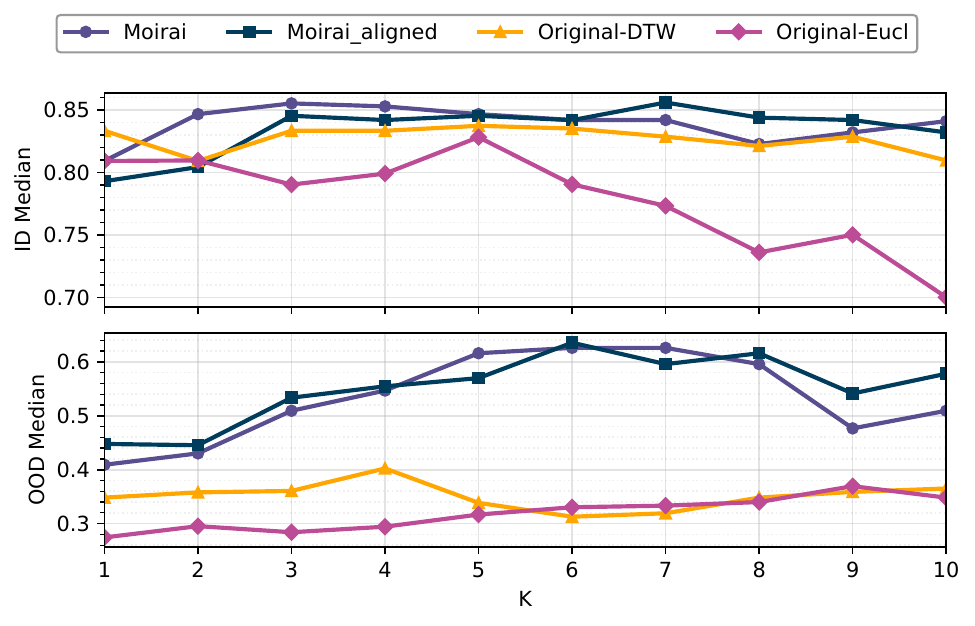}
    \vspace{-0.7cm}
    \caption{Moirai}
  \end{subfigure}

  \vspace{0.2em}

  \begin{subfigure}[t]{0.4\textwidth}
    \centering
    \includegraphics[width=\linewidth]{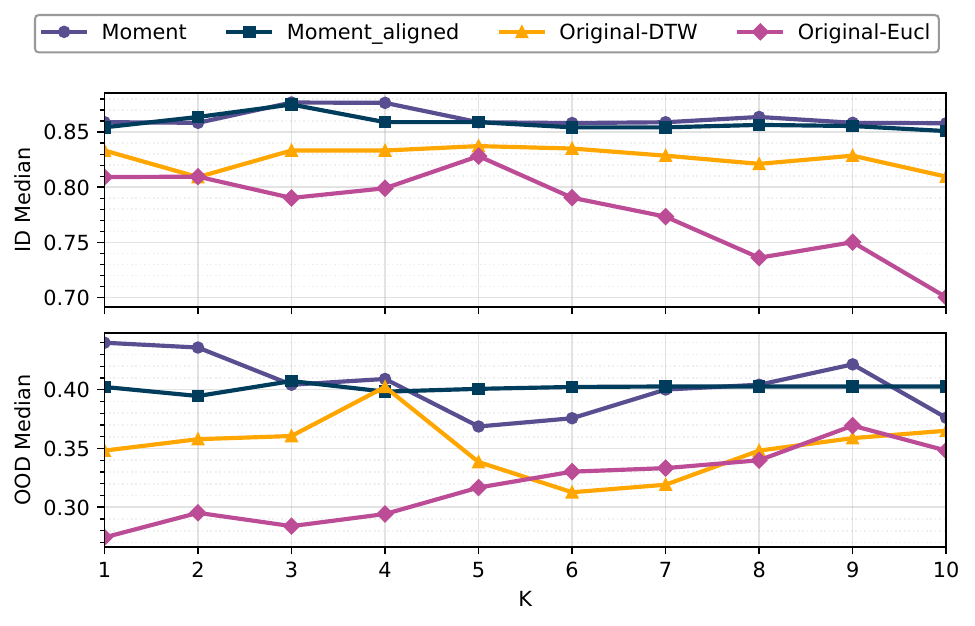}
    \vspace{-0.7cm}
    \caption{Moment}
  \end{subfigure}
  \hfill
  \begin{subfigure}[t]{0.4\textwidth}
    \centering
    \includegraphics[width=\linewidth]{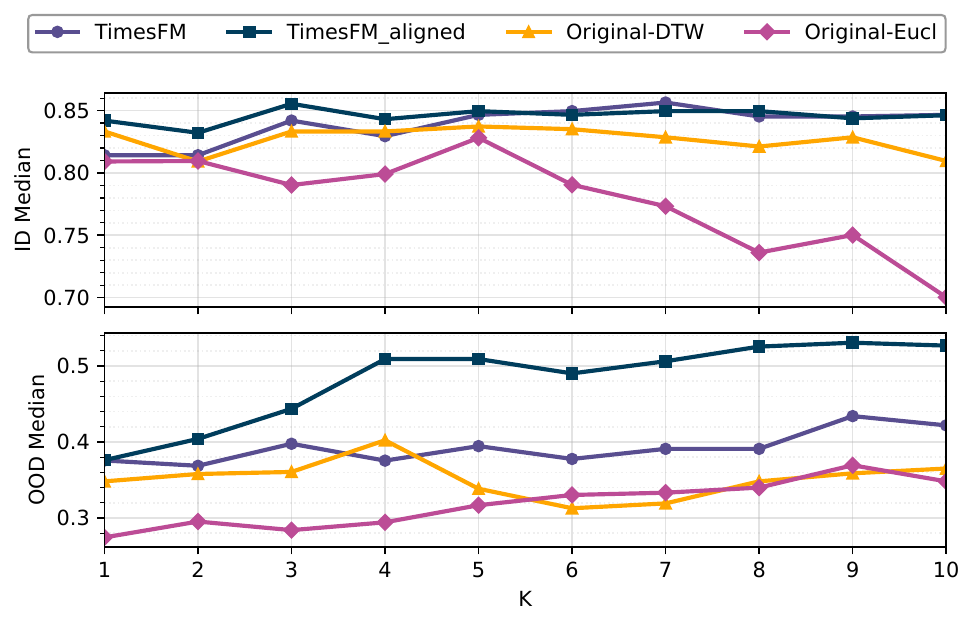}
    \vspace{-0.7cm}
    \caption{TimesFM}
  \end{subfigure}

  \vspace{0.2em}

  \begin{subfigure}[t]{0.4\textwidth}
    \centering
    \includegraphics[width=\linewidth]{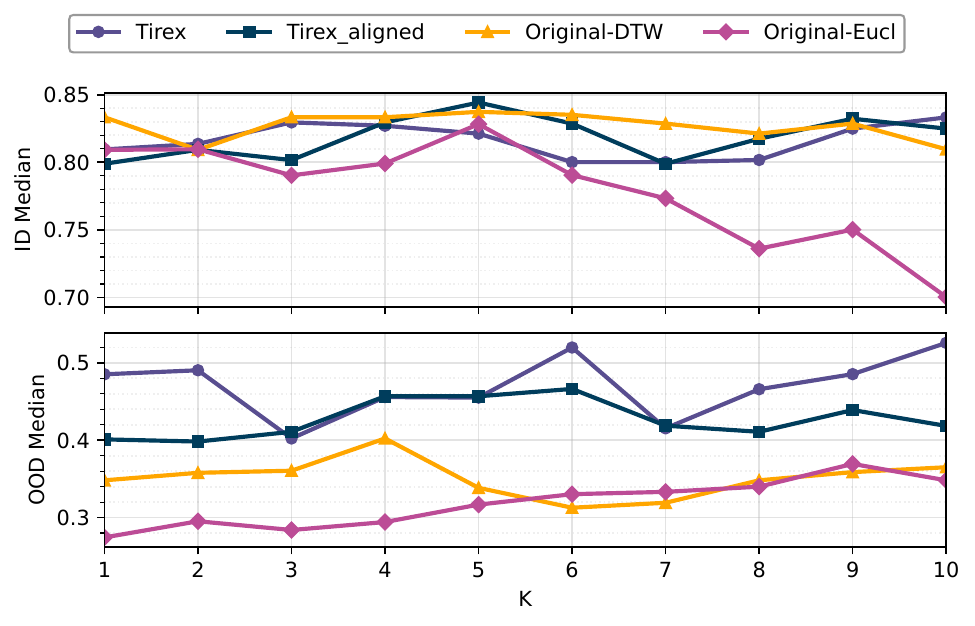}
    \vspace{-0.7cm}
    \caption{Tirex}
  \end{subfigure}
  \hfill
  \begin{subfigure}[t]{0.4\textwidth}
    \centering
    \includegraphics[width=\linewidth]{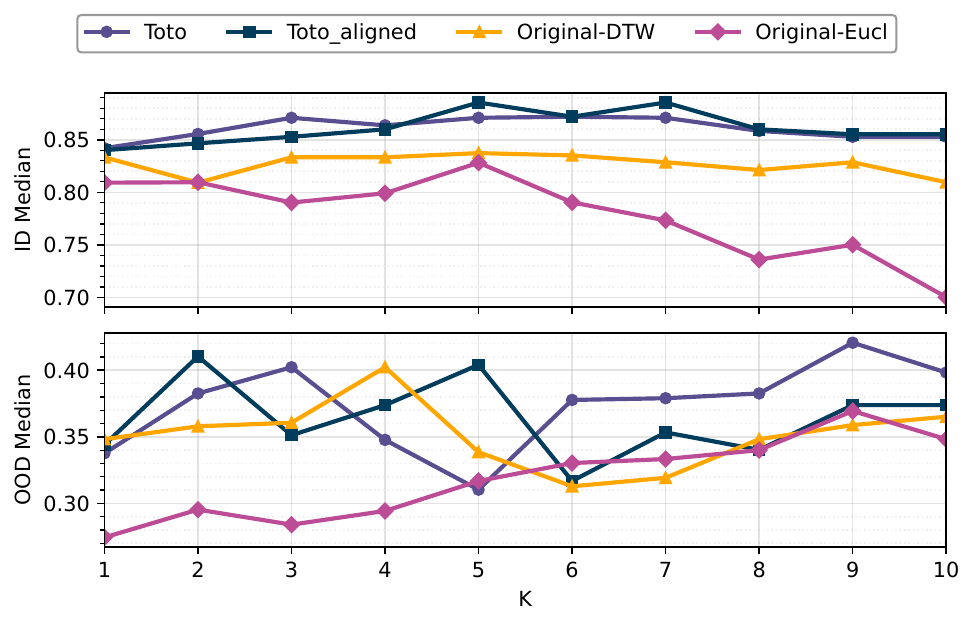}
    \vspace{-0.7cm}
    \caption{Toto}
  \end{subfigure}

  \caption{
    Sensitivity analysis with respect to parameter $k$ across eight forecasting models.
    Each panel reports the model-specific response under identical evaluation conditions,
    allowing direct visual comparison of trend, variability, and stability.
  }
  \label{fig:variation-k-all-models}
\end{figure}

\section{Experimental Setup and Benchmark Details}
\label{app:experimental_details}

\subsection{Dataset Details}
\label{app:datasets}

For the evaluation of RAMSAD, we considered the TSB-AD~\cite{liu2024elephant} benchmark, which includes a comprehensive set of 870 curated univariate time series from 9 domains, which are summarized in Table \ref{tab:datasets}. To ensure a fair comparison, we used the same train/test splits as recent model selection benchmarks~\cite{liu2025tsb}. The splits are the same for both ID and OOD scenarios. However, in the latter, for a given time series, automated solutions can only consider training series that do not belong to its domain. For example, for a test series belonging to the \textit{Medical} category, automated solutions can only consider training series that belongs to the other $8$ domains (e.g., a meta-learning approach can be trained only on this reduced training subset).

\begin{table}[tb]
\caption{Dataset Domains.}
\label{tab:datasets}
\footnotesize
\centering
\resizebox{0.5\textwidth}{!}{%
\begin{tabular}{lccc}
\toprule
\textbf{Domain} &
  \textbf{Total} &
  \textbf{Training} & \textbf{Test}\\ \midrule
Environment  & 20 &  14 & 6  \\ 
Facility  & 143 &  102 & 41  \\ 
Finance  & 20 &  14 & 6  \\ 
Human Activity & 58 & 41 & 17 \\ 
Medical & 147 & 105 & 42 \\
Sensor & 44 & 32 & 12 \\ 
Synthetic & 122 & 87 & 35 \\ 
Traffic & 6 & 4 & 2 \\ 
WebService & 310 & 220 & 90 \\ \midrule
\textbf{Total}  & \textbf{870}  & \textbf{619}  &  \textbf{251}\\  \bottomrule

\end{tabular}%
}
\end{table}

\subsection{List of Candidate Detectors}
\label{app:detectors}

Table \ref{tab:tsad_detectors} lists all the $32$ anomaly detection approaches considered in our experimental evaluation. As it can be seen, the pool of available models covers all the different 
categories described in the literature review, and aligns with the one used in recent model selection benchmarks \cite{liu2025tsb}.

\begin{table}[tb]
\caption{List of anomaly detection approaches included in our experimental evaluation.}
\label{tab:tsad_detectors}
\footnotesize
\resizebox{\textwidth}{!}{%
\begin{tabular}{l|l|p{9cm}}
\toprule
\textbf{Approach} &
  \textbf{Category} &
  \textbf{Description} \\ \midrule
Sub-LOF \cite{breunig2000lof} & Distance-based
   &
  \multirow{2}{*}{Identifies density-based local outliers.} \\ 
LOF \cite{breunig2000lof} & Distance-based
   &
   \\ \midrule
MatrixProfile \cite{yeh2016matrix} & Distance-based
   &
  Computes the distance between all subsequences. \\ \midrule
KShapeAD \cite{paparrizos2015k} & Distance-based
   &
  Identifies the normal pattern  based on the k-Shape clustering algorithm and computes distance-based anomaly scores. \\ \midrule
SAND \cite{boniol2021sand} &  Distance-based
   &
  Incrementally updates a summary of normal patterns using subsequences from  batches and evaluates anomaly scores based on distance. \\ \midrule
Series2Graph \cite{boniol2020series2graph} & Density-based
   &
   Transforms subsequences into nodes and their transition patterns into edges. Anomalies are   identified based on irregular transition patterns. \\ \midrule
SR \cite{ren2019time} & Distance-based
   &
  Computes anomaly scores as the relative difference between saliency map values and their moving averages. \\ \midrule
Sub-KNN \cite{ramaswamy2000efficient} & Distance-based
   &
  Leverages the distance to the k-th nearest neighbor. \\ \midrule
KMeansAD \cite{yairi2001fault} & Distance-based
   &
  Measures the distance to the centroid of  assigned cluster. \\ \midrule
  Sub-IForest \cite{liu2008isolation} &  Density-based
   &
  \multirow{2}{9cm}{Isolates anomalies using random partitioning of the data space.} \\ 
IForest \cite{liu2008isolation} & Density-based
   &
   \\ \midrule
Sub-PCA \cite{aggarwal2017introduction} & Density-based
   &
  Projects data to a lower-dimensional   hyperplane, with significant deviation from this plane indicating anomalies. \\ \midrule
Sub-HBOS \cite{goldstein2012histogram} & Density-based
   &
  Constructs a histogram for the data and uses the inverse of the height of the bin as the anomaly score of the data point. \\ \midrule
Sub-OCSVM \cite{scholkopf1999support} & Density-based
   &
  Fits the dataset to find the normal   data's boundary by maximizing the margin between the origin and the normal samples. \\ \midrule
Sub-MCD \cite{rousseeuw1999fast} & Density-based
   &
  Leverages minimum covariance determinant \\ \midrule
POLY \cite{li2007unifying} & Density-based
   &
  Uses local polynomial approximation. \\ \midrule

AutoEncoder \cite{sakurada2014anomaly} & Prediction-based
   &
  Leverages reconstruction errors through encoding-decoding. \\ \midrule
CNN \cite{munir2018deepant} & Prediction-based
   &
 Leverages prediction errors through CNN. \\ \midrule
LSTMAD \cite{malhotra2015long} & Prediction-based
   &
 Leverages prediction errors through LSTM.  \\ \midrule
TranAD \cite{tuli2022tranad} & Prediction-based
   &
  Leverages self-conditioning and adversarial   training. \\ \midrule
AnomalyTransformer \cite{xu2021anomaly} & Prediction-based
   &
  Leverages Anomaly-Attention mechanism to compute association discrepancies. \\ \midrule
OmniAnomaly \cite{su2019robust} & Prediction-based
   &
  Leverages a Stochastic recurrent neural network. \\ \midrule
USAD \cite{audibert2020usad} & Prediction-based
   &
  Leverages adversely trained autoencoders. \\ \midrule
Donut \cite{xu2018unsupervised} & Prediction-based
   &
  Based on Variational Auto-Encoders (VAE).  \\ \midrule
TimesNet \cite{wu2022timesnet} & Prediction-based
   &
  Based on temporal 2D-variation modeling. \\ \midrule
FITS \cite{xu2023fits} & Prediction-based
   &
   Interpolates time series in the frequency domain. \\ \midrule
TimesFM \cite{das2023decoder} & Prediction-based
   &
  Pre-trains a decoder-only attention model with input patching. \\ \midrule
MOMENT (ZS) \cite{goswami2024moment} & Prediction-based
   &
  Uses a Zero Shot, pre-trained transformer model. \\ \midrule
MOMENT (FT) \cite{goswami2024moment} & Prediction-based
   &
    Uses a Fine-Tuned, pre-trained transformer model. \\ \midrule
OFA \cite{zhou2023one} & Foundation model
   &
  Based on fine-tuned GPT-2 model. \\ \midrule
Lag-Llama \cite{rasul2023lag} & Foundation model
   &
  Leverages a decoder-only transformer architecture. \\ \midrule
Chronos \cite{ansari2024chronos} & Foundation model
   &
  Tokenizes time series values using scaling and quantization. \\ \bottomrule

\end{tabular}%
}
\end{table}

\subsection{Automated TSAD Baselines}
\label{app:baselines}

Table~\ref{tab:autotsad} lists the $25$ automated solutions that we compared as baselines. More specifically, we include $2$ theoretical baselines, i.e., selecting always the best detector (Oracle) and selecting the best detector basing on a labeled portion of the target series. We also include $3$ simple baselines (i.e., Random selection, Global best, and Foundation model). In addition to these baselines, we included a comprehensive set of state-of-the-art automated solutions, including $8$ model selection approaches, $4$ internal selection methods, $5$ ensembles strategies and $3$ model generation approaches.

\begin{table}[tb]
\caption{List of automated solutions compared with RAMSAD.}
\label{tab:autotsad}
\footnotesize
\resizebox{\textwidth}{!}{%
\begin{tabular}{l|l|p{9cm}}
\toprule
\textbf{Approach} &
  \textbf{Category} &
  \textbf{Description} \\ \midrule
Oracle  & Theoretical baseline &  Selects the single best anomaly detector for each series  \\ \midrule
Supervised Selection (SS) & Theoretical baseline &{ Oracle-style upper bound: selects the best detector using labels of a portion of the target series, unavailable to the other methods} \\ \midrule
Global Best (GB) & Baseline & Global best: selects the detector with the best overall accuracy on the entire benchmark. \\ \midrule
Random  & Baseline  &  Randomly selects one of the detectors for each series.  \\ \midrule
Foundation model (FM) \cite{das2023decoder}  & Baseline &  Foundation model pre-trained on large scale datasets. \\ \midrule
SATZilla \cite{xu2008satzilla} & Model Selection  &  Builds per-instance algorithm portfolios using empirical hardness models to predict performance and select the best SAT solver for each instance.  \\ \midrule
ARGOSMART \cite{nikolic2013simple}  & Model Selection  &  Uses 1NN on dataset meta-features to select the model that performed best on the most similar training dataset. \\ \midrule
ISAC \cite{kadioglu2010isac}  & Model Selection  &  Clusters datasets via meta-features and selects the best-performing model from the nearest cluster to the new dataset.\\ \midrule
MetaOD \cite{zhao2021automatic}  & Model Selection  &  Leverages collaborative filtering and matrix factorization on dataset meta-features to predict and rank candidate models for a new dataset.  \\ \midrule
MSAD \cite{sylligardos2023choose}  & Model Selection  & Uses classification to map datasets to the best-performing anomaly detection model among a set of candidates.  \\ \midrule
UReg \cite{navarro2023meta} & Model Selection  &  Learns a regression from dataset meta-features to latent SVD factors to predict model performance for novel datasets.  \\ \midrule
CFact \cite{navarro2023meta} & Model Selection  &  Clusters configurations in latent space and trains cluster-specific regressors to predict performance and select the best model for known datasets.  \\ \midrule
Model Centrality (MC) \cite{lin2020infogan}  & Model Selection  &  Computes model centrality via Kendall’s  distances between anomaly score rankings to select models closest to their nearest neighbors. \\ \midrule
Clustering Quality (CQ) \cite{nguyen2016evaluation}  & Internal  & Applies cluster validation metrics on thresholded anomaly scores to select models with better separation between normal/anomalous traces.  \\ \midrule
Unsupervised evaluation curves (UEC) \cite{goix2016evaluate}  & Internal &  Evaluates anomaly detectors using unsupervised mass-volume and excess-mass curves instead of labeled ROC/PR-based metrics.  \\ \midrule
Synthetic \cite{chatterjee2022mospat}  & Internal  &  Injects artificial anomalies into simulated time series and selects the model with the best detection performance.\\ \midrule
TSAdams \cite{goswami2022unsupervised} & Internal  &  Aggregates surrogate metric rankings of anomaly detectors using Kemeny/Borda methods to produce a more robust final model ranking.  \\ \midrule
AutoTSAD \cite{schmidl2024autotsad} & Ensemble & Generates synthetic anomalous time series to optimize detector configurations and ensemble their ranked anomaly scores into a final prediction. \\ \midrule
Outlier Ensemble (OE) \cite{aggarwal2015theoretical}  & Ensemble  &  Uses an ensemble of all available detectors. \\ \midrule
SELECT \cite{rayana2016less} & Ensemble  &  Applies unsupervised ensemble selection to choose a subset of detectors using vertical and horizontal consensus strategies for final aggregation.  \\ \midrule
Iterative Outlier Ensemble (IOE) \cite{ma2023need}  & Ensemble  &  Iteratively builds a pseudo–ground truth by selecting and refining anomaly score consensus to produce a final aggregated ensemble score.  \\ \midrule
HITS \cite{ma2023need}  & Ensemble  & Recursively computes hub and centralities over anomaly score relations to rank and select more reliable models. \\ \midrule
AutoOD-A \cite{cao2023autood} & Model Generation  &  Iteratively refines pseudo-labels via voting and learns to prune weak detectors using logistic regression and multi-view updates. \\ \midrule
AutoOD-C \cite{cao2023autood} & Model Generation  &  Iteratively cleans noisy pseudo-labels by tracking early training losses to remove likely mislabeled instances and refine anomaly detection.  \\ \midrule
UADB \cite{ye2023uadb} & Model Generation  &  Applies knowledge distillation to transfer multiple anomaly detectors’ knowledge into a booster model that adapts and improves detection across datasets.  \\  \bottomrule

\end{tabular}%
}
\end{table}

\subsection{Hardware Details}
\label{app:hardware}

All experiments were performed on a machine running Linux kernel 5.4.0-212-generic. The node is equipped with two Intel(R) Xeon(R) Gold 6242R CPUs, providing 80 logical CPU cores across 2 sockets with 2 threads per core. The system has 503 GiB of main memory and an 8 GiB swap partition. GPU-accelerated experiments used a single NVIDIA Quadro RTX 6000 with 23 GiB of onboard memory.

\section{Additional Ablation Studies}
\label{app:ablations}

\subsection{Effect of Reducing the Training Size}
\label{app:reduced_training}

In this part, we evaluate the performance of RAMSAD when reducing the size of the training set. 
Figure~\ref{fig:reduced_sizes} shows the average VUS-PR achieved by RAMSAD with training size varying from $5\%$ to $100\%$ of the original set, comparing it with the performance of the best model selector (i.e., SATZilla) in ID and OOD scenarios. 
By considering the former scenario, we can see that RAMSAD can outperform SATZilla even with just the $80$\% of the training set, and that it achieves comparable scores even with sizes of $20\%$, $40\%$, and $60\%$. 
These results underscore the ability of RAMSAD of retrieving meaningful examples from the knowledge base, enabling effective model selection. 
This aspect is even more evident in the OOD scenario. 
As we can see, RAMSAD outperforms SATZilla even when using just the $10$\% of the original training set (i.e.,$62$ series).

\subsection{Effect of Reducing the Pool Size of Detectors}
\label{app:reduced_detector_pool}

Figure \ref{fig:reduced_sizes} also reports the performance of RAMSAD when randomly varying the pool of available detectors in ID and OOD scenarios. Each value reported represents the average across $10$ repetitions to take into account the variability due to the random sampling.  In both scenarios, we can see that RAMSAD is able to {match or} outperform SATZilla even when considering only half of the original pool (i.e., $16$ detectors); {note that SATZilla is shown with its performance on the full detector pool, so this comparison is conservative for RAMSAD}, highlighting that, by retrieving meaningful examples, it allows to select an effective detector even in this more challenging scenario.

\section{Additional Discussions}

\subsection{Limitations}
\label{app:limitations}

Although RAMSAD achieves strong performance for time-series anomaly detector selection, it still has several limitations.

\textbf{Segmentation-induced information loss.}
To take advantage of TSFM representations, RAMSAD partitions each time series into fixed-length segments before retrieval. While this makes heterogeneous series compatible with the encoder, it may also discard information carried by longer-range temporal dependencies and by the global structure of the original series. As a result, similarity computed from segment-level representations may not fully reflect similarity at the full-series level. More generally, the segmentation strategy itself is fixed rather than optimized, and we do not yet study how to choose segment boundaries or segment lengths so as to minimize retrieval distortion. {Adaptive segmentation and encoders with longer contexts are natural directions to reduce this loss. Relatedly, we do not resample series, and therefore make no claim of invariance to the sampling rate.}

\textbf{Remaining gap to the oracle.}
RAMSAD provides a strong non-parametric baseline, but it does not fully close the gap to the theoretical oracle in either the ID or the OOD setting. As shown in Figure~\ref{fig:id_vs_ood}, there is still room for improvement in how retrieved neighbors are identified and how detector recommendations are transferred from them. Reducing this gap remains an important direction for future work.

\textbf{Restriction to univariate time series.}
Our current study focuses only on univariate time series. Although the overall retrieval-based framework could be extended to multivariate settings, such an extension is not explored here. Adapting RAMSAD to multivariate anomaly detection, including the design of suitable segment representations and similarity spaces, is therefore left for future work.

\textbf{Temporal concept drift and streaming.}
Our experiments concern domain shift between static series: the query comes from a domain that is absent from the knowledge base. We do not evaluate temporal concept drift within a domain or a stream, nor non-stationarity within a single long series, such as an anomalous pattern that becomes normal over time. Nothing in the method precludes these cases, since RAMSAD retrieves per segment and aggregates segment matches into a series ranking, so later segments of an evolving series are embedded and retrieved afresh and the recommendation can change as they accumulate. We present this as a direction for future work, together with a streaming evaluation protocol, and not as a claim supported by our experiments.

\textbf{Labeled knowledge base and queries far from every covered domain.}
RAMSAD requires a labeled knowledge base on which every candidate detector has been evaluated, the same supervision used by trained meta-learning selectors, and it does not apply where no labels exist at all. Our leave-one-domain-out protocol evaluates entirely unseen domains: no series, performance profile or description from the target domain enters the knowledge base or the alignment training. Missing descriptions are not a failure mode either, since a query never supplies one. What our experiments do not isolate is a query that is far from every indexed domain in representation space. Since what matters for selection is proximity in that space rather than shared domain membership, such a query may retrieve weakly related neighbors, and characterizing when the knowledge-base coverage is insufficient, and how to build a knowledge base that avoids it, is left for future work.

\subsection{Broader Impacts}
\label{app:broader_impacts}

Improved model selection for time-series anomaly detection can support more reliable monitoring systems in domains such as healthcare, manufacturing, finance, energy, cybersecurity, and scientific monitoring. 
By reducing the need to run large detector pools, RAMSAD may also improve the efficiency and accessibility of anomaly-detection pipelines. 
At the same time, anomaly-detection systems can produce false alarms or missed detections, which may have negative consequences in high-stakes deployments if used without appropriate human oversight. 
The method could also be applied in monitoring or surveillance contexts, where privacy and governance considerations depend on the deployment setting. 
We therefore view RAMSAD as a decision-support component rather than a replacement for domain-specific validation and responsible deployment practices.

\end{document}